\documentclass{article}
\usepackage{array}
\newcolumntype{P}[1]{>{\raggedright\arraybackslash}p{#1}}
\usepackage[T1]{fontenc}     
\usepackage{ragged2e}        
\usepackage[protrusion=true,expansion=false]{microtype}  
\usepackage{graphicx}
\usepackage{multirow}
\usepackage{caption}
\usepackage{float}
\usepackage{booktabs}
\usepackage{multicol}
\usepackage{makecell}
\usepackage{tablefootnote}
\usepackage{textgreek} 
\usepackage{tabularx}        
\usepackage{tikz}

\usepackage[table,xcdraw]{xcolor}
\usepackage{amsmath,amssymb,amsfonts}
\usepackage{mathtools} 
\usepackage{amsthm}
\usepackage{mathrsfs}
\usepackage{algorithm}
\usepackage{algpseudocode}  
\usepackage{comment}

\usepackage{textcomp}
\usepackage{pifont}
\usepackage{setspace,lineno}
\usepackage{listings}
\usepackage{soul}            
\sethlcolor{yellow}          
\usepackage[title]{appendix}
\usepackage[authoryear,round]{natbib}
\usepackage{manyfoot}
\usepackage{cite}
\usepackage{xcolor}

\definecolor{customgreen}{HTML}{00B050}
\definecolor{captionblue}{HTML}{0070C0}  
\newcommand{\cmark}{\textcolor{green}{\ding{51}}} 
\newcommand{\xmark}{\textcolor{red}{\ding{55}}} 
\DeclareCaptionLabelFormat{boldblue}{\textbf{\textcolor{captionblue}{#1 #2}}}
\usepackage{titlesec}
\titleformat{\section}
  {\normalfont\Large\bfseries\color{captionblue}}{\thesection}{1em}{}
\titleformat{\subsection}
  {\normalfont\large\bfseries\color{captionblue}}{\thesubsection}{1em}{}
\titleformat{\subsubsection}
  {\normalfont\normalsize\bfseries\color{captionblue}}{\thesubsubsection}{1em}{}
\usepackage{geometry}
\definecolor{orcidgreen}{HTML}{A6CE39}

\newcommand{\orcidicon}[1]{%
  \textsuperscript{%
    \href{https://orcid.org/#1}{%
      \begin{tikzpicture}[baseline=-0.1em]
        \fill[orcidgreen] (0,0) circle (1.5ex);
        \node[white, scale=0.8, font=\bfseries\sffamily] at (0,0) {iD};
      \end{tikzpicture}%
    }%
  }%
}

\usepackage{amsmath} 
\allowdisplaybreaks[4]
\usepackage[protrusion=true,expansion=false]{microtype}
\usepackage{hyperref}
\hypersetup{
    colorlinks=true,
    linkcolor=blue,
    citecolor=blue,
    urlcolor=blue
}
\title{Multi-Modal Semantic Segmentation of Electrolyzer Components for Sustainable Hydrogen Technologies: A Dual-Branch Deep Learning Approach}

\author{
Wasimul Karim\textsuperscript{1, 2}\orcidicon{0009-0008-9886-8110}, 
Nur Mohammad Fahad\textsuperscript{1,3,*}\orcidicon{0009-0005-3577-2523}, 
Abdul Hasib Siddique\textsuperscript{2, 4}\orcidicon{0000-0002-5122-7040}, \\
Md Rafiqul Islam\textsuperscript{4}\orcidicon{0000-0001-7209-3881}, 
Hooman Mehdizadeh-Rad\textsuperscript{4}\orcidicon{0000-0002-5849-9292},
Asif Karim\textsuperscript{4}\orcidicon{0000-0001-8532-6816},
Sami Azam\textsuperscript{4,*}\orcidicon{0000-0001-7572-9750} \\
\small
\textsuperscript{1}Applied Artificial Intelligence and INtelligent Systems (AAIINS) Laboratory, Dhaka 1217, Bangladesh \\
\small
\textsuperscript{2}Department of Computer Science and Engineering, University of Scholars, 40 Kemal Ataturk Avenue, Dhaka 1213, Bangladesh \\
\small
\textsuperscript{3}School of Engineering and Energy, Murdoch University, Murdoch, WA 6150, Australia\\
\small
\textsuperscript{4}Faculty of Science and Technology, Charles Darwin University, Casuarina, NT 0909, Australia \\
\small 
\textsuperscript{*}Corresponding Authors: \href{mailto:35996218@murdoch.student.edu.au}{35996218@murdoch.student.edu.au}, \href{mailto:sami.azam@cdu.edu.au}{sami.azam@cdu.edu.au}
}

\date{} 
\begin{document}


\justifying
\twocolumn[
\maketitle
\begin{abstract}
\noindent Accurate segmentation of electrolyzer materials is essential for automated disassembly, sustainable recycling, and circular manufacturing in hydrogen technologies. However, this task is challenging due to strong visual similarity between materials, spectral overlap, irregular shapes, and severe class imbalance. To address these challenges, we propose an AI-driven dual-branch framework, Hyperspectral-RGB Electrolyzer Materials Network (HREM-Net), that combines hyperspectral imaging (HSI) and RGB images for electrolyzer material segmentation. We implemented several innovative modules, including Efficient Channel Attention, Coordinate Attention, Mobile Inverted Bottleneck blocks, and Atrous Spatial Pyramid Pooling to capture spectral and spatial features from HSI, and RGB images. With an adaptive gated cross-modal fusion module and composite loss function, HREM-Net achieves a mean class accuracy of 91.66\% and a mean Intersection over Union (mIoU) of 0.82 on the Electrolyzers-HSI dataset, outperforming baseline segmentation models. Cross-dataset validation on the PCB-Vision dataset demonstrates strong generalization with 96.91\% accuracy and 0.93 mIoU. This work poses its potential as an industrial application to improve electrolyzer efficiency, thereby improving the predictive maintenance of hydrogen production.
\end{abstract}

\vspace{0.5em}
\vspace{1em}
]

\section{Introduction}\label{sec1}
Recycling and reusing recovered materials play an important role in reducing the environmental impacts of hydrogen technologies \citep{lotrivc2021life, valente2019end, ferriz2019end}. Sustainable recycling requires fast and accurate automated industrial material detection systems; however, reliable segmentation remains challenging because industrial materials have irregular shapes, complex surfaces, overlapping structures, and uneven lighting, while RGB images capture surface details, yet often fail to distinguish visually similar or time-varying materials \citep{jiang2026smfps, arbash2025electrolyzers}. Automated disassembly of end-of-life electrolyzer materials combined with effective recycling strategies can reduce environmental impacts and material costs, contributing to the sustainability of hydrogen technologies \citep{kaiser2025recycling, uekert2024electrolyzer}. In automated electrolyzer material recycling, segmentation maps can directly be used to guide robotic sorting, material separation, and downstream recycling decisions \citep{bacchin2024wastegan, yousif2025leveraging, de2021domestic, koskinopoulou2021robotic} to avoid cross-material contamination and ensure high-purity recovery. As hydrogen technologies develop and electrolyzer recycling becomes more important for sustainability \citep{uekert2024electrolyzer, axt2025towards, ishtiaq2025survey}, accurate material segmentation requires combining complementary sensing modalities to enable reliable identification and recovery of key visually similar materials \citep{konstantinidis2023multi, xu2025complementary}. In this context, Hyperspectral Imaging (HSI) captures wavelength-dependent reflectance characteristics that reveal material composition \citep{heist20185d}, while RGB imaging provides detailed spatial structure, texture information, and precise boundary localization \citep{jiang2025global}. Such heterogeneous cross-modal information is difficult to leverage effectively using single modality approaches \citep{wang2025cross}. This highlights the need for a dual-branch architecture that separately learns spectral features from hyperspectral data and spatial structure from RGB images, while adaptively combining both through attention-based fusion. Using dedicated encoders, dynamic modality weighting, and an appropriate loss function helps to address both local details and global semantic information, class imbalance, and achieve reliable material segmentation \citep{wang2025image, fu2025cafnet, ma2021loss, yeung2022unified}. 

Despite the growing interest in hydrogen technologies, studies on electrolyzer material segmentation remain limited. Existing methods have progressed from RGB-based single-modality models to ensemble and transformer-based approaches, where early works relied on RGB images for waste and material recognition, with detection-based models \citep{rahmatulloh2025wasteinnet, lu2022using} learning spatial patterns and weakly supervised \citep{yudin2024hierarchical} and data-centric methods \citep{iliushina2025data} improving scalability. Similarly, Convolutional Neural Network (CNN)-based approaches for fuel cell segmentation \citep{tang2022deep} and SOFC microstructure analysis \citep{tan2026image} focused on spatial features and phase boundaries. More recent studies explored hyperspectral imaging (HSI), where spectral information supports better material separation, as shown by U-Net variants with spectral encoders \citep{picon2024hyperspectral}. In parallel, ensemble and transformer-based models \citep{wang2025cross, jafar2026ensemble} improve feature combination for segmentation. Although these studies \citep{wang2025cross, rahmatulloh2025wasteinnet, lu2022using, yudin2024hierarchical, iliushina2025data, tang2022deep, tan2026image, picon2024hyperspectral, jafar2026ensemble, senanayake2025automated} report strong results, several limitations remain. Most methods rely on a single modality, using grayscale microscopy \citep{tang2022deep, tan2026image}, hyperspectral data alone \citep{picon2024hyperspectral}, or RGB images \citep{wang2025cross, rahmatulloh2025wasteinnet, lu2022using, yudin2024hierarchical, iliushina2025data, jafar2026ensemble, senanayake2025automated}, without combining spectral and spatial information, which limits discrimination of visually similar materials such as Steel-Black and Steel-Grey. Hyperspectral methods \citep{picon2024hyperspectral} provide material-specific signatures, yet lack a dedicated RGB branch and proper fusion, reducing boundary quality. In addition, most architectures \citep{wang2025cross, rahmatulloh2025wasteinnet, lu2022using, tang2022deep, tan2026image, picon2024hyperspectral, jafar2026ensemble, senanayake2025automated} do not use attention mechanisms, limiting feature focus and causing confusion between similar materials. Finally, none of the methods use specialized loss functions, making it difficult to handle severe class imbalance and boundary errors in the electrolyzer material dataset. In industrial recycling environments, additional challenges may arise from surface oxidation, contamination, thermal aging, or material mixing after long-term operation. Such conditions can influence both visual appearance and spectral response, increasing the difficulty of reliable material recognition across different electrolyzer components.

\begin{figure*}[ht!]
\centering
\includegraphics[scale=0.67]{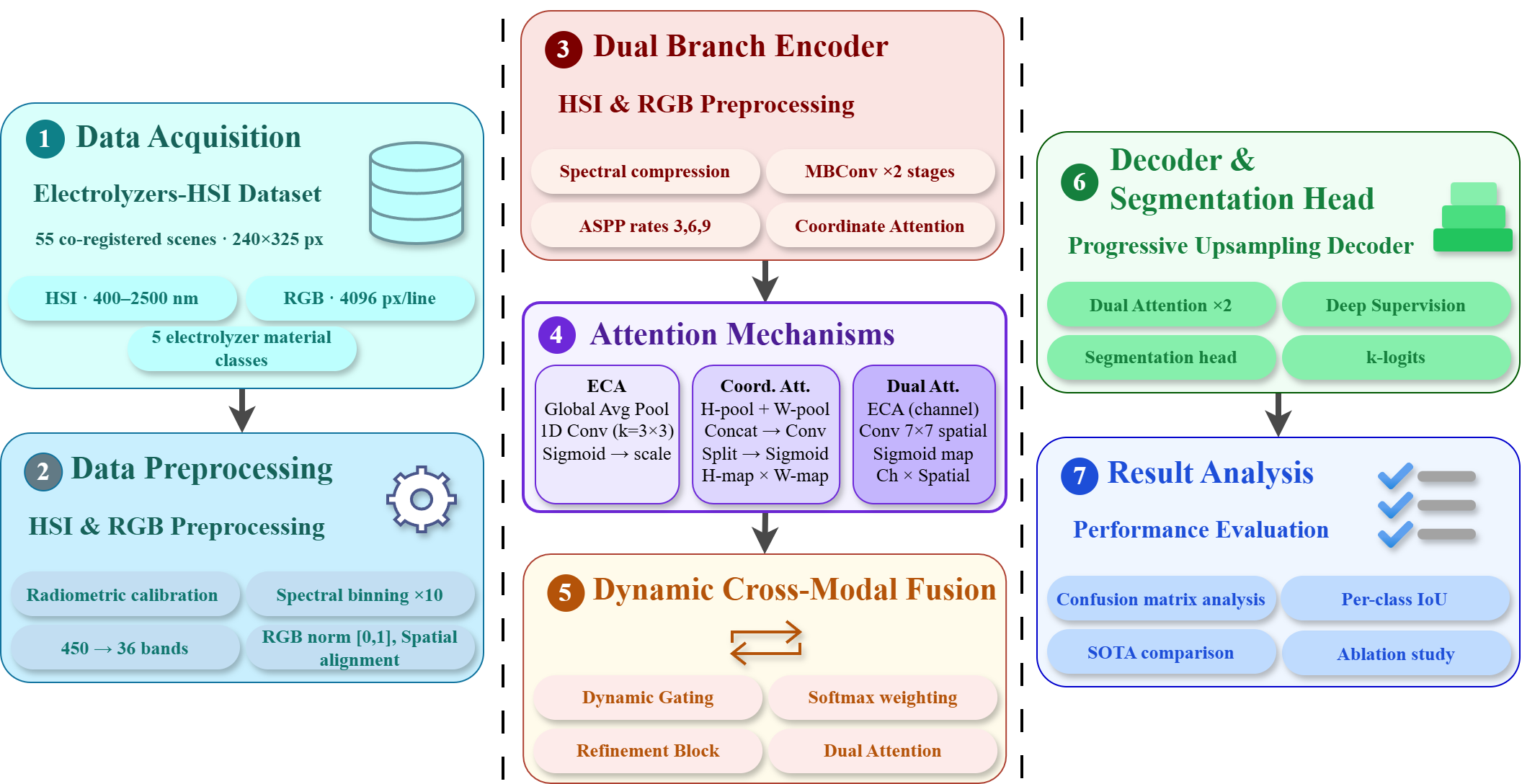}
\caption{Workflow diagram of the proposed study.}
\label{fig:1}
\end{figure*}

HREM-Net addresses these shortcomings by introducing a balanced multimodal framework, preserving superior segmentation performance. It adopts a dual-branch architecture that processes hyperspectral and RGB data in parallel, allowing each branch to focus on different aspects of material understanding.  Figure \ref{fig:1} depicts the workflow diagram of this proposed study. The contributions of this work can be summarized as follows:

\begin{itemize}
    \item Introduces HREM-Net, the first dual-branch HSI-RGB fusion architecture that jointly employs spectral compression and dynamic cross-modal gating to achieve accurate and modality-aware electrolyzer material segmentation.
    \item Proposes an attention-based fusion method that uses Efficient Channel Attention, Coordinate Attention, and cross-modal fusion with dynamic gating to refine cross-modal feature interaction.
    \item Mobile inverted bottleneck blocks (MBConv) with squeeze-excitation (SE) and Atrous Spatial Pyramid Pooling (ASPP) are employed on the extracted features to capture spatial details and multi-scale context. 
    \item Employs a composite optimization strategy based on a multi-component loss that integrates PolyLoss, Tversky loss, and auxiliary deep supervision. This formulation improves optimization stability, mitigates class imbalance, and enhances segmentation accuracy near material boundaries.
    \item Demonstrates SOTA performance showing consistent improvements over existing methods across different material classes, including those with highly imbalanced data.
\end{itemize}

The remainder of the paper is organized as follows:

Section \ref{sec2} reviews related work on material segmentation in materials science and waste management applications. Section \ref{sec3} introduces the proposed HREM-Net architecture, describing all the attention modules, cross-modal fusion strategy, and the multi-component loss function. Section \ref{sec4} explains the datasets and the preprocessing for model training. Section \ref{sec5} presents the experimental results, and the paper's core contribution, limitations, and future research plan are discussed in Section \ref{sec6}. Finally, Section \ref{sec7} concludes the paper.

\section{Related Works}
\label{sec2}

Growing resource consumption and global waste are major global challenges of our time \citep{wurster2022modelling}. Recent studies on waste and material segmentation and detection are described in this section, with a particular emphasis on RGB-based single-modality segmentation models, HSI approaches, and ensemble and transformer-based frameworks, as these have been the recent focus for handling complex material discrimination and generalization challenges across diverse domains.

\subsection{RGB-Based Segmentation Models}
Single modality RGB-based models are commonly used as baselines for waste detection and segmentation using standard object detection and semantic segmentation architectures. Rahmatulloh et al.~\citep{rahmatulloh2025wasteinnet} proposed WasteInNet, a YOLOv7-tiny-based model for real-time multi-class indoor waste detection, trained on the WasteIn dataset with 1,210 images across 7 categories, achieving a precision of 0.801, recall of 0.873, and mean Average Precision at IoU threshold 0.5 (mAP@0.5) of 0.868 using data augmentation. Lu et al.~\citep{lu2022using} developed a DeepLabv3+ model with an Xception65 backbone and Atrous Spatial Pyramid Pooling (ASPP) for mixed construction waste recognition, trained on 5,022 images from 9 categories collected in Hong Kong, reporting a mean Intersection over Union (mIoU) of 0.555 with Materials in Context Database (MINC) pre-training and Output Stride=8. Yudin et al.~\citep{yudin2024hierarchical} introduced H-YC, a hierarchical model combining YOLOX-m detection with ConvNeXt-based Class Activation Map segmentation to avoid pixel-level annotation, achieving an mAP@0.5 of 59.6\% and a weakly supervised segmentation mIoU of 0.6588 on the WaRP dataset with 10,406 images across 28 categories. Iliushina et al.~\citep{iliushina2025data} used YOLOv8m with a data-centric pipeline including pseudo-labeling, object-based augmentation, and class balancing on a 22-class Municipal Solid Waste (MSW) dataset, achieving an mAP@0.5 of 70\% through two-stage pre-training and fine-tuning.

In another study, Tan et al.~\citep{tan2026image} proposed an inspired three-dimensional CNN U-Net model that removes all max pooling and upsampling layers to keep full voxel-level spatial resolution. The model performs three-phase Solid Oxide Fuel Cell (SOFC) anode segmentation across three Focused Ion Beam Scanning Electron Microscopy (FIB-SEM) datasets, and achieves average diameter accuracies of more than 90\%. However, these RGB-only systems have a common limitation. They cannot reliably distinguish between materials that look visually similar, yet have different spectral properties. These studies didn't incorporate advanced cross-modal fusion architectures, dynamic gating for modality weighting, and comprehensive attention modules to address the visual similarity issue and severe class imbalance.

\begin{table*}[ht!]
\centering
\scriptsize
\caption{Comparative analysis of existing waste and material segmentation methods and the proposed HREM-Net, highlighting modality, architecture, dataset, key limitations, and distinguishing contributions.}
\label{tab:comparison}
\begin{tabular}{
>{\raggedright\arraybackslash}p{2.5cm}
>{\centering\arraybackslash}p{1.25cm}
>{\centering\arraybackslash}p{2cm}
>{\centering\arraybackslash}p{2cm}
>{\centering\arraybackslash}p{1.5cm}
>{\centering\arraybackslash}p{1.5cm}
>{\centering\arraybackslash}p{5cm}
}
\toprule
\textbf{Study} & \textbf{Modality} & \textbf{Architecture} & \textbf{Fusion Strategy} & \textbf{Attention Mechanism} & \textbf{Imbalance Loss} & \textbf{Limitations} \\
\midrule
Wang et al.~\citep{wang2025cross} & RGB & Swin + Twins + K-Net & probability weighting & None & None & No class imbalance handling; single modality \\
Rahmatulloh et al.~\citep{rahmatulloh2025wasteinnet} & RGB & YOLOv7-tiny & None & None & None & Poor class accuracy; single modality \\
Lu et al.~\citep{lu2022using} & RGB & DeepLabv3+ & None & ASPP only & None & Material ambiguity; no class imbalance handling\\
Yudin et al.~\citep{yudin2024hierarchical} & RGB & YOLOX + ConvNeXt & Hierarchical & Basic CAM & None & RGB-only limits discrimination\\
Iliushina et al.~\citep{iliushina2025data} & RGB & YOLOv8m & None & None & None & Ambiguity; no class imbalance handling\\
Tang et al.~\citep{tang2022deep} & Micro-CT & U-ResNet & None & None & None & No spectral integration; standard CE loss only\\
Tan et al.~\citep{tan2026image} & FIB-SEM & 3D CNN U-Net & None & None & None & Overestimation; sensitive to noise \\
Picon et al.~\citep{picon2024hyperspectral} & HSI & U-Net, DINOv2 & None & None & None & Very small dataset; no cross-modal fusion \\
Jafar et al.~\citep{jafar2026ensemble} & RGB & U-Net + FPN & Ensemble Average & None & None & Struggles with deformed/cluttered objects \\
Senanayake et al.~\citep{senanayake2025automated} & RGB & VLPart (Mask R-CNN) & None & None & None & RGB-only limits discrimination; No class imbalance handling \\
\midrule
\textbf{HREM-Net (Ours)} & \textbf{HSI + RGB} & \textbf{Dual-Branch Encoder-Decoder} & \textbf{Cross-modal + Dynamic Gating} & \textbf{ECA + CA + Cross-Attention} & \textbf{PolyLoss + Tversky + Auxiliary} & \_ \\
\bottomrule
\end{tabular}
\end{table*}

\subsection{Instance-Level, Transformer-Based, and Hyperspectral Approaches}
Fine-grained instance segmentation methods have extended waste recognition from simple category-level detection to more detailed material understanding, as discussed in recent studies \citep{senanayake2025automated, jafar2026ensemble}. Senanayake et al. \citep{senanayake2025automated} used fine-tuned VLPart, a vision language instance segmentation model built on Mask Region-based CNN (Mask R-CNN) and Contrastive Language Image Pretraining (CLIP) text embeddings. On a custom dataset of 1,000 images with 14 subcategories, the Swin Base backbone achieved a mAP@0.5 of 39.6, outperforming Residual Network 50 and Residual Network Next 101 (ResNext-101) by 8.60\% and 3.64\%, respectively. Jafar et al. \citep{jafar2026ensemble} evaluated seven models on the ZeroWaste-f industrial waste dataset and proposed EL-4, an ensemble of U-Net and Feature Pyramid Network (FPN) with a shared EfficientNet-B4 encoder, which achieved an Intersection over Union (IoU) of 0.8306, compared with 0.8065 for U-Net alone.

Transformer-based and ensemble models have shown improved performance in construction material segmentation by capturing richer contextual features than traditional convolution-based models \citep{wang2025cross, tang2022deep}. Wang et al. \citep{wang2025cross} evaluated several transformer architectures and proposed a two-stage ensemble strategy with probability-based weighting, achieving a mIoU of 0.826 and a pixel accuracy of 90.30\% on a custom dataset. Similarly, Tang et al. \citep{tang2022deep} proposed U ResNet, which combines a U-Net style encoder-decoder design with Residual Network blocks, reporting 92\% pixel accuracy and a mIoU of 0.64 on Proton Exchange Membrane Fuel Cell (PEMFC) micro-CT segmentation data. HSI-based methods have also shown strong potential for material segmentation, as they use spectral reflectance information that is not available in RGB images. Picon et al. \citep{picon2024hyperspectral} evaluated several models, including one-dimensional and two-dimensional encoder-decoder networks, a U-Net style model, and DINOv2-based variants, on the Tecnalia dataset for Waste Electrical and Electronic Equipment (WEEE). The U-Net style model with seven selected spectral bands achieved the best result, with a mIoU of 0.74 and an accuracy of 95\%, outperforming both the full 76-band setup (0.73 mIoU) and the RGB-only baseline (0.64 mIoU).

The proposed HREM-Net addresses the key limitations observed across the existing studies \citep{wang2025cross, rahmatulloh2025wasteinnet, lu2022using, yudin2024hierarchical, iliushina2025data, tang2022deep, tan2026image, picon2024hyperspectral, jafar2026ensemble, senanayake2025automated}, as presented in Table~\ref{tab:comparison}. In contrast to RGB-only models \citep{wang2025cross, rahmatulloh2025wasteinnet, lu2022using, yudin2024hierarchical, iliushina2025data, senanayake2025automated, jafar2026ensemble, tang2022deep}, HREM-Net integrates dedicated Spectral and RGB encoder branches, enabling reliable discrimination of visually similar materials such as Steel-Black and Steel-Grey that RGB alone cannot resolve. Different from the hyperspectral-only approach of Picon et al.~\citep{picon2024hyperspectral}, which foregoes spatial RGB detail, HREM-Net performs full HSI-RGB dual-branch fusion to jointly process spectral features and spatial boundary precision. No other study adopted HSI and RGB processing simultaneously to improve the segmentation performance. Furthermore, in contrast to all reviewed methods, our innovative loss function has established effective training and deals with class-imbalance and boundary segmentation errors, which traditional loss functions fail to address. 

\section{Methodology}
\label{sec3}
\begin{figure*}[ht!]
\centering
\includegraphics[scale=0.44]{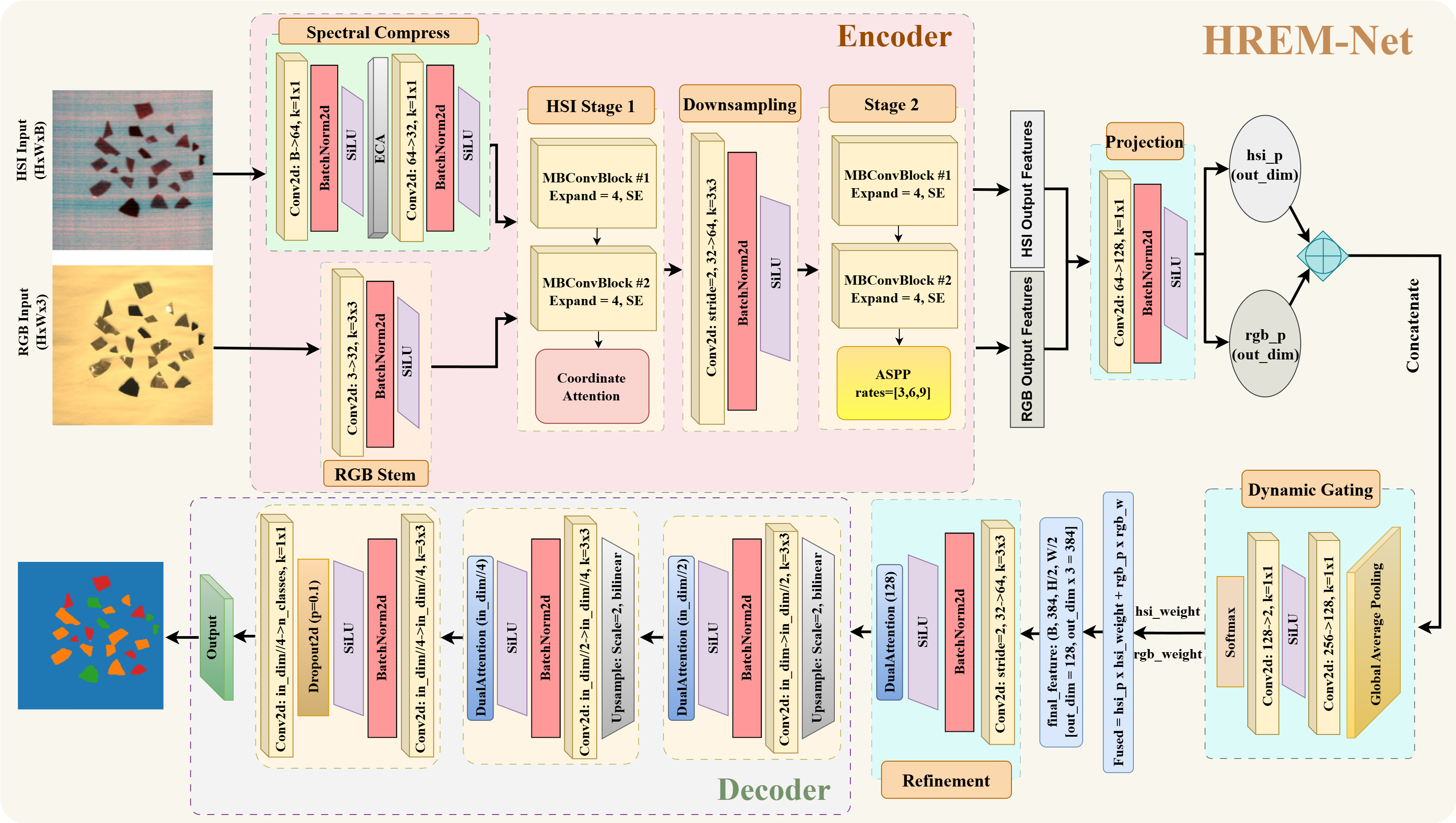}
\caption{Overview of the proposed HREM-Net architecture for multimodal electrolyzer material segmentation. Hyperspectral and RGB inputs are processed through separate encoders to extract spectral and spatial features, which are then adaptively fused using attention-guided cross-modal fusion with dynamic gating and decoded to produce accurate pixel-level segmentation.}
\label{fig:hremnet}
\end{figure*}

The overall methodology of the proposed HREM-Net is illustrated in Figure \ref{fig:hremnet}. This section explains the complete methodology, including problem formulation, dual-branch encoding, attention mechanisms, cross-modal fusion, and the multi-component loss function.

\subsection{Problem Formulation}
As hydrogen production capacity expands worldwide, understanding and managing the environmental impacts associated with electrolyzer materials recycling has become increasingly important \citep{khan2024strategies, iyer2024life}. We formulate electrolyzer material recognition as a multimodal semantic segmentation problem. Semantic segmentation divides an image into meaningful regions by assigning a class label to every pixel, enabling accurate object identification and localization \citep{franke2025vision, zhou2026generative}. Given a dataset $\mathcal{D} = \{(X_i, Y_i)\}_{i=1}^{N}$, each sample consists of two complementary inputs: a hyperspectral image $x^{(h)}_i \in \mathbb{R}^{B \times H \times W}$ and a spatially aligned RGB image $x^{(r)}_i \in \mathbb{R}^{3 \times H \times W}$. The target label $Y_i \in \{1,\dots,C\}^{H \times W}$ represents the pixel-wise ground truth annotation for $C$ different electrolyzer material classes, where $B$ denotes the number of spectral bands.

Each modality provides different yet useful information. Hyperspectral images capture wavelength-dependent reflectance characteristics that are important for distinguishing materials with very similar visual appearance. In contrast, RGB images provide detailed spatial structure, texture, and edge information that support accurate boundary localization. Our objective is to learn a segmentation function $\mathcal{F}_\theta$ that can effectively combine these two inputs to produce a dense per-pixel class score map, given by $\hat{Y} = \mathcal{F}_\theta(x^{(h)}_i, x^{(r)}_i)$.

The final semantic label at each pixel position $(u,v)$ is determined by selecting the class with the highest predicted score. This formulation allows the model to jointly use spectral information and spatial context when assigning material labels. This study extebded RGB only model by using the hyperspectral data, which adds high spectral dimensionality and strong correlations between bands, making efficient feature compression necessary while still preserving material-specific cues. Furthermore, many electrolyzer components have similar shapes and visual patterns, meaning that spatial information alone is often not sufficient. The presence of strong class imbalance further increases the difficulty of the learning task.

To handle these challenges, we propose a dual-branch segmentation model that processes hyperspectral and RGB inputs separately while allowing adaptive interaction between the two modalities. The model parameters $\theta$ are optimized by minimizing a segmentation loss over the training dataset, defined as
$\theta^{*} = \arg\min_{\theta} \frac{1}{N} \sum_{i=1}^{N} \mathcal{L}\big(\mathcal{F}_{\theta}(x_i^{(h)}, x_i^{(r)}), Y_i\big)$, where $\mathcal{L}(\cdot)$ is a composite loss function designed to address class imbalance and maintain accurate material boundaries.

\subsection{Dual-Branch Encoder Architecture}

The proposed HREM-Net is built upon a dual-branch encoder: a hyperspectral encoder $\mathcal{E}_{h}$ designed to compress and refine spectral features, and an RGB encoder $\mathcal{E}_{r}$ designed to preserve spatial layout and texture details. The two feature streams are combined only after they have been transformed into stable and meaningful representations. This separation provides advantages in maintaining physical hyperspectral signatures and geometric structure without interference from feature redundancy.

\subsubsection{Hyperspectral Encoder}

The hyperspectral encoder converts high-dimensional hyperspectral cube into a compact feature representation that preserves material-discriminative spectral information while also capturing local spatial patterns required for dense segmentation. The hyperspectral input consists of multiple spectral bands stacked along the channel dimension with spatial resolution $(H,W)$, where each pixel contains a full spectral signature. Processing such high-dimensional data directly with standard convolutions is inefficient because many spectral bands are highly correlated and redundant \citep{zhuang2020hyperspectral, morales2021hyperspectral}.

To address this issue, the encoder begins with an explicit spectral compression stage. This compression module uses two successive $1\times1$ convolutions combined with channel attention to perform efficient spectral mixing at each pixel without spatial pooling. The first projection layer reduces the spectral dimensionality while preserving important information through convolution, batch normalization, and nonlinear activation. The first projection stage is defined as equation \ref{eq2}:

\begin{equation}
\mathbf{F}{h}^{(1)} = \sigma\left(\mathrm{BN}\left(\mathbf{W}_{h1} * \mathbf{X}^{h}\right)\right),
\label{eq2}
\end{equation}

where $\mathbf{W}_{h1}$ maps spectral channels from $B$ to $2D$, $\sigma(\cdot)$ denotes a nonlinear activation function, $\mathrm{BN}(\cdot)$ denotes batch normalization, and $*$ represents a $1\times1$ point-wise convolution. Here, $D$ denotes the target embedding dimension of the hyperspectral encoder. The intermediate features are then recalibrated using Efficient Channel Attention (ECA) \citep{wang2020eca} and projected to the final embedding dimension as described in equation \ref{eq3}:

\begin{equation}
\mathbf{F}{h}^{(2)} = \sigma\left(\mathrm{BN}\left(\mathbf{W}_{h2} * \mathrm{ECA}\left(\mathbf{F}_{h}^{(1)}\right)\right)\right),
\label{eq3}
\end{equation}

where $\mathbf{W}_{h2}$ maps feature channels from $2D$ to $D$. This design allows the network to first analyze spectral importance through attention and then compress the features to the final embedding dimension. As a result, the encoder can emphasize informative spectral responses before the final projection stage.

The ECA mechanism is embedded within the spectral compression pipeline to adaptively adjust channel responses. Instead of using heavy fully connected layers, ECA models local channel interactions through a lightweight one-dimensional convolution. Given the intermediate feature map $\mathbf{F}_{h}^{(1)} \in \mathbb{R}^{2D \times H \times W}$, the global channel descriptor is computed as defined in equation \ref{eq4}:

\begin{equation}
z_c = \frac{1}{HW}\sum_{u=1}^{H}\sum_{v=1}^{W}\mathbf{F}_{h,c}^{(1)}(u,v),
\label{eq4}
\end{equation}

where $z_c$ represents the global descriptor of channel $c$, and $\mathbf{F}_{h,c}^{(1)}(u,v)$ denotes the activation value at spatial location $(u,v)$ of channel $c$.

Channel-wise attention weights are then generated through a one-dimensional convolution:

\begin{equation}
a_c = \sigma\left(\sum_{k=-\lfloor K/2 \rfloor}^{\lfloor K/2 \rfloor} w_k , z_{c+k}\right),
\label{eq5}
\end{equation}

Here in equation \ref{eq5}, $a_c$ denotes the attention weight for channel $c$, $\sigma(\cdot)$ is the sigmoid activation function, $w_k$ represents the learnable convolution kernel weights, and $K$ is the kernel size of the one-dimensional convolution that determines the local channel interaction range. These attention weights are applied to recalibrate the feature channels, allowing the network to emphasize more informative spectral responses while suppressing less useful ones.

After spectral compression and channel attention, the encoder extracts spatial features using stacked Mobile Inverted Bottleneck Convolution (MBConv) blocks \citep{sandler2018mobilenetv2}. Each MBConv block follows an inverted bottleneck design, where channels are first expanded, spatial filtering is applied using depthwise convolution, and then projected back to the target dimension. This design supports efficient spatial feature learning with low computational cost, which is important for high-resolution hyperspectral data. To improve feature selectivity, a Squeeze-and-Excitation (SE) \citep{shu2022expansion} module is included, which recalibrates channel-wise features using global spatial context. This helps the network identify the most important channels for representing material characteristics. The channel attention vector is computed as shown in equation \ref{eq8}:

\begin{equation}
\mathbf{s} = \sigma_{\text{sig}}\left(\mathbf{W}_2 \cdot \delta\left(\mathbf{W}_1
\cdot \text{GAP}(\mathbf{V})\right)\right),
\label{eq8}
\end{equation}

where $\text{GAP}(\cdot)$ denotes global average pooling, $\mathbf{W}_1$ and $\mathbf{W}2$ are the squeeze and excitation weight matrices that project the channel dimension from $C \rightarrow C/r$ and $C/r \rightarrow C$, respectively, $\delta(\cdot)$ represents the SiLU activation function, $\sigma{\text{sig}}(\cdot)$ denotes the sigmoid function, and $\mathbf{s}$ is the learned channel attention vector.

The generated attention vector is then applied to the feature map through element-wise channel scaling, allowing the network to emphasize informative channels while suppressing less useful ones. After channel recalibration, the features are projected back to the desired output dimension using a point-wise convolution followed by batch normalization. This projection step reduces the expanded channels produced by the MBConv block and prepares the features for the residual connection. Let $\mathbf{F}_{\text{in}}$ represent the input feature map of an MBConv block. When the input and output channel dimensions are identical, and the spatial resolution is preserved (i.e., stride equals one), a residual connection is employed to facilitate effective feature propagation. In this case, stochastic regularization is introduced in the residual branch through spatial dropout, which is applied to the projected features before they are added to the input. This helps reduce overfitting and improves generalization during training. In contrast, when the channel dimensions differ or spatial downsampling is performed, the residual connection is omitted, and only the transformed features are propagated forward.

To further preserve spatial information beyond spectral compression, Coordinate Attention \citep{hou2021coordinate} is applied after the initial MBConv feature extraction stage. In contrast to standard channel attention mechanisms that collapse spatial information completely, Coordinate Attention decomposes global pooling into two one-dimensional encoding operations along the height and width directions. This design enables the network to capture long-range spatial dependencies while still preserving precise positional information. The directional feature aggregation is computed as represented in equation \ref{eq12}:

\begin{equation}
\mathbf{g}_{h}(i) = \frac{1}{W}\sum_{j=1}^{W}\mathbf{F}_{h}^{\text{stage1}}(i,j), \quad
\mathbf{g}_{w}(j) = \frac{1}{H}\sum_{i=1}^{H}\mathbf{F}_{h}^{\text{stage1}}(i,j),
\label{eq12}
\end{equation}

where $\mathbf{F}_{h}^{\text{stage1}}$ denotes the feature map after MBConv processing. The directional descriptors are then concatenated along the spatial dimension and processed through a shared bottleneck transformation shown in equation \ref{eq13}:

\begin{equation}
\mathbf{h} = \sigma\left(\mathrm{BN}\left(\mathbf{W}_{\text{reduce}} * [\mathbf{g}{h}; \mathbf{g}_{w}]\right)\right)
\label{eq13}
\end{equation}

The resulting tensor $\mathbf{h}$ is then split along the spatial dimension into $\mathbf{h}{split}^{(h)}$ and $\mathbf{h}{split}^{(w)}$, which correspond to height-aligned and width-aligned feature embeddings, respectively. In this step, concatenation is performed along the spatial dimension rather than the channel dimension, and $\mathbf{W}_{\text{reduce}}$ reduces the channel dimension from $C$ to $C/r$, with the reduction ratio set to $r=8$. The processed features are then projected separately to generate directional attention maps, as presented in equation \ref{eq14}:

\begin{equation}
\mathbf{A}_{h} = \sigma_{\text{sig}}\left(\mathbf{W}_{h} * \mathbf{h}_{split}^{(h)}\right), \quad
\mathbf{A}_{w} = \sigma_{\text{sig}}\left(\mathbf{W}_{w} * \mathbf{h}_{split}^{(w)}\right)
\label{eq14}
\end{equation}

Finally, these directional attention maps are applied to the feature representation through element-wise multiplication along the height and width directions. This operation enables the network to preserve spatial structure while emphasizing informative regions, which is particularly beneficial for maintaining material boundaries and expanded spectral patterns. This design offers strong local feature modeling while keeping the number of parameters low, making it suitable for close-range scenes where fine spatial details and edges need to be preserved. Electrolyzer components appear at different scales across images. For example, mesh regions often form thin and expanded structures, while steel plates can cover large continuous areas. To support scale-robust feature learning under these conditions, we apply Atrous Spatial Pyramid Pooling (ASPP) \citep{lian2021cascaded}:

\begin{equation}
\mathbf{F}_{\text{ASPP}} = \text{Proj}\left(\bigoplus_{r \in \mathcal{R}} \sigma\left(\mathrm{BN}\left(\mathrm{Conv}_{r}(\mathbf{F}_{h}')\right)\right) \oplus \text{GAP}(\mathbf{F}_{h}')\right),
\label{eq16}
\end{equation}

In equation \ref{eq16}, $\mathbf{F}_{h}'$ denotes the input feature map, and $\mathcal{R}=\{1, 3, 6, 9\}$ defines the dilation rates used in $\mathrm{Conv}_{r}(\cdot)$. Here, $r=1$ corresponds to standard convolution, while larger values indicate atrous convolutions with increased receptive fields. The operator $\text{GAP}$ represents global average pooling followed by a $1 \times 1$ convolution and upsampling to match the original spatial size. The symbol $\oplus$ denotes channel-wise concatenation, and $\text{Proj}$ is a projection layer that merges all branches into a compact feature representation. This multi-scale context encoding helps reduce over-segmentation in highly textured regions and lowers confusion between small minority classes and dominating classes. Algorithm \ref{alg:spectral_encoder} presents the detailed workflow of the spectral encoder.

\begin{algorithm}[ht!]
\caption{Spectral Encoder for HSI}
\begin{algorithmic}[1]
\Require HSI input $\mathbf{X}_{HSI} \in \mathbb{R}^{B \times H \times W}$, base dimension $D_{base}$
\Ensure Encoded features $\mathbf{F}_{enc} \in \mathbb{R}^{2D_{base} \times H/2 \times W/2}$
\State \textbf{// Spectral Compression with Attention}
\State $\mathbf{F} \gets \text{SiLU}(\text{BN}(\text{Conv}_{1 \times 1}(\mathbf{X}_{HSI}, 2D_{base})))$ \Comment{Initial projection}
\State $\mathbf{F} \gets \text{EfficientChannelAttention}(\mathbf{F})$ \Comment{Channel attention}
\State $\mathbf{F} \gets \text{SiLU}(\text{BN}(\text{Conv}_{1 \times 1}(\mathbf{F}, D_{base})))$ \Comment{Compress channels}
\State \textbf{// Stage 1: Spatial Feature Extraction}
\State $\mathbf{F} \gets \text{MBConvBlock}(\mathbf{F}, D_{base}, \text{expand}=4)$ \Comment{Inverted bottleneck 1}
\State $\mathbf{F} \gets \text{MBConvBlock}(\mathbf{F}, D_{base}, \text{expand}=4)$ \Comment{Inverted bottleneck 2}
\State $\mathbf{F} \gets \text{CoordinateAttention}(\mathbf{F})$ \Comment{Spatial attention}
\State \textbf{// Downsampling}
\State $\mathbf{F} \gets \text{SiLU}(\text{BN}(\text{Conv}_{3 \times 3, s=2}(\mathbf{F}, 2D_{base})))$ 
\State \textbf{// Stage 2: Multi-Scale Context}
\State $\mathbf{F} \gets \text{MBConvBlock}(\mathbf{F}, 2D_{base}, \text{expand}=4)$ 
\State $\mathbf{F} \gets \text{MBConvBlock}(\mathbf{F}, 2D_{base}, \text{expand}=4)$ 
\State $\mathbf{F}_{enc} \gets \text{ASPP}(\mathbf{F}, 2D_{base}, \text{rates}=[3,6,9])$ \Comment{Multi-scale aggregation}
\State \Return $\mathbf{F}_{enc}$
\end{algorithmic}
\label{alg:spectral_encoder}
\end{algorithm}

\subsubsection{RGB Encoder}
The RGB encoder is responsible for extracting spatial details \citep{jiang2018rednet}. Although hyperspectral data provides strong material information, the RGB modality serves a key role in highlighting boundaries, capturing geometric structure, and resolving fine edges in close-range images. This is particularly important for thin structures, mesh boundaries, and electrode structures, where pixel-level accuracy depends heavily on clear boundary definition.

The RGB input contains three intensity values that describe the visible appearance of the scene. The encoder begins with a stem convolution that converts raw pixel intensities into low-level feature responses as highlighted in equation \ref{eq18}:

\begin{equation}
\mathbf{F}{r}^{(0)} = \sigma\left(\mathrm{BN}\left(\mathbf{W}_{r0} * \mathbf{X}^{r}\right)\right)
\label{eq18}
\end{equation}

where $\mathbf{W}_{r0}$ represents the convolution weights of the stem layer. Similar to the hyperspectral branch, MBConv blocks are used to build an efficient feature hierarchy by learning texture, edge direction, and boundary consistency. Downsampling uses strided convolutions to increase the receptive field while preserving structural features. To retain spatial location awareness, Coordinate Attention is added to the RGB branch. It aggregates information separately along height and width instead of both together, allowing the network to keep positional information while capturing long-range spatial dependencies. These directional features are then concatenated and passed through a shared bottleneck transformation with a reduction ratio of $r=8$, as defined in equation \ref{eq19}:

\begin{equation}
\mathbf{h} = \sigma\left(\mathrm{BN}\left(\mathbf{W}_{\text{reduce}} * [\mathbf{g}_{h}, \mathbf{g}_{w}]\right)\right),
\label{eq19}
\end{equation}

where $\mathbf{W}_{\text{reduce}}: C \rightarrow C/r$ performs channel reduction and height-wise and width-wise pooled descriptors $\mathbf{g}{h}$ and $\mathbf{g}_{w}$ are computed from the RGB feature map. The tensor $\mathbf{h}$ is split along the spatial dimension into $\mathbf{h}{split}^{(h)}$ and $\mathbf{h}{split}^{(w)}$, corresponding to height- and width-aligned feature embeddings. As shown in equation \ref{eq20}, the transformed features are then split and separately projected to generate directional attention maps:

\begin{equation}
\mathbf{A}_{h} = \sigma_{\text{sig}}\left(\mathbf{W}_{h} * \mathbf{h}_{split}^{(h)}\right), \quad
\mathbf{A}_{w} = \sigma_{\text{sig}}\left(\mathbf{W}_{w} * \mathbf{h}_{split}^{(w)}\right),
\label{eq20}
\end{equation}

These attention maps are then applied to the RGB feature representation through element-wise modulation along the height and width directions, emphasizing spatially informative regions while preserving fine structural details, which is particularly useful for detecting thin expanded structures such as meshes and electrode boundaries. Algorithm \ref{alg:dual_branch_net} presents the overall workflow of the proposed Dual-Branch Network.

\begin{algorithm*}[ht!]
\caption{Dual-Branch Network}
\begin{algorithmic}[1]
\Require HSI image $\mathbf{I}_{HSI} \in \mathbb{R}^{B \times H \times W}$, RGB image $\mathbf{I}_{RGB} \in \mathbb{R}^{3 \times H \times W}$
\Ensure Segmentation map $\mathbf{S} \in \mathbb{R}^{K \times H \times W}$, auxiliary output $\mathbf{S}_{aux}$, gate weights $(\alpha, \beta)$
\State \textbf{// Dual-Branch Feature Extraction}
\State $\mathbf{F}_{HSI} \gets \text{SpectralEncoder}(\mathbf{I}_{HSI}, D_{base})$ 
\State $\mathbf{F}_{RGB} \gets \text{RGBEncoder}(\mathbf{I}_{RGB}, D_{base})$ 
\State \textbf{// Cross-Modal Fusion}
\State $\mathbf{F}_{fused}, (\alpha, \beta) \gets \text{CrossModalFusion}(\mathbf{F}_{HSI},  \mathbf{F}_{RGB}, D_{fusion})$ 
\State \textbf{// Progressive Upsampling Decoder}
\State $\mathbf{F}_{up1} \gets \text{DualAttention}(\text{SiLU}(\text{BN}(\text{Upsample}_{2\times}(\text{Conv}_{3 \times 3}(\mathbf{F}_{fused}, D_{fusion}/2)))))$
\If{training}
    \State $\mathbf{S}_{aux} \gets \text{BilinearInterpolate}(\text{Conv}_{1 \times 1}(\mathbf{F}_{up1}, K), (H, W))$ \Comment{Deep supervision}
\EndIf
\State $\mathbf{F}_{up2} \gets \text{DualAttention}(\text{SiLU}(\text{BN}(\text{Upsample}_{2\times}(\text{Conv}_{3 \times 3}(\mathbf{F}_{up1}, D_{fusion}/4)))))$
\State \textbf{// Segmentation Head}
\State $\mathbf{F}_{seg} \gets \text{Dropout}(\text{SiLU}(\text{BN}(\text{Conv}_{3 \times 3}(\mathbf{F}_{up2}, D_{fusion}/4))))$
\State $\mathbf{S} \gets \text{BilinearInterpolate}(\text{Conv}_{1 \times 1}(\mathbf{F}_{seg}, K), (H, W))$ \Comment{Final prediction}
\State \Return $\mathbf{S}, \mathbf{S}_{aux}, (\alpha, \beta)$
\end{algorithmic}
\label{alg:dual_branch_net}
\end{algorithm*}

\subsection{Attention Mechanisms}

Instead of using a single attention module, we combine three attention mechanisms to address different challenges in multimodal material segmentation. ECA \citep{wang2020eca} selects important channels with low computational cost and highlights useful spectral channels for distinguishing visually similar materials, which is important for hyperspectral encoders. Coordinate Attention \citep{hou2021coordinate} captures long-range dependencies while preserving spatial coordinate information, which helps the RGB encoder maintain accurate boundaries and structure, especially for thin components. Finally, Dual Attention (DA) \citep{fu2019dual} performs channel reweighting and spatial refinement in a two-step process. In the first step, ECA is applied to the input feature tensor $\mathbf{F}$ to adaptively recalibrate channel responses as $\mathbf{F}_{\text{ch}} = \mathrm{ECA}(\mathbf{F})$. This operation produces channel-refined features, denoted as $\mathbf{F}_{\text{ch}}$, where informative channels are emphasized and less relevant responses are suppressed.

In the second step, spatial attention is generated by applying a large-kernel convolution ($7\times7$) to the channel-refined features $\mathbf{F}_{\text{ch}}$, followed by a sigmoid activation function. This operation produces a spatial attention map $\mathbf{M}_{s}$ that highlights important spatial regions while reducing background responses. The final refined representation is then obtained by element-wise multiplying the channel-refined features $\mathbf{F}_{\text{ch}}$ with the spatial attention map $\mathbf{M}_{s}$. This two-stage refinement first emphasizes discriminative channels and then suppresses background-related metallic texture responses while strengthening object-consistent regions, which directly improves segmentation smoothness and boundary sharpness.

\begin{figure}[ht!]
\centering
\includegraphics[scale=0.24]{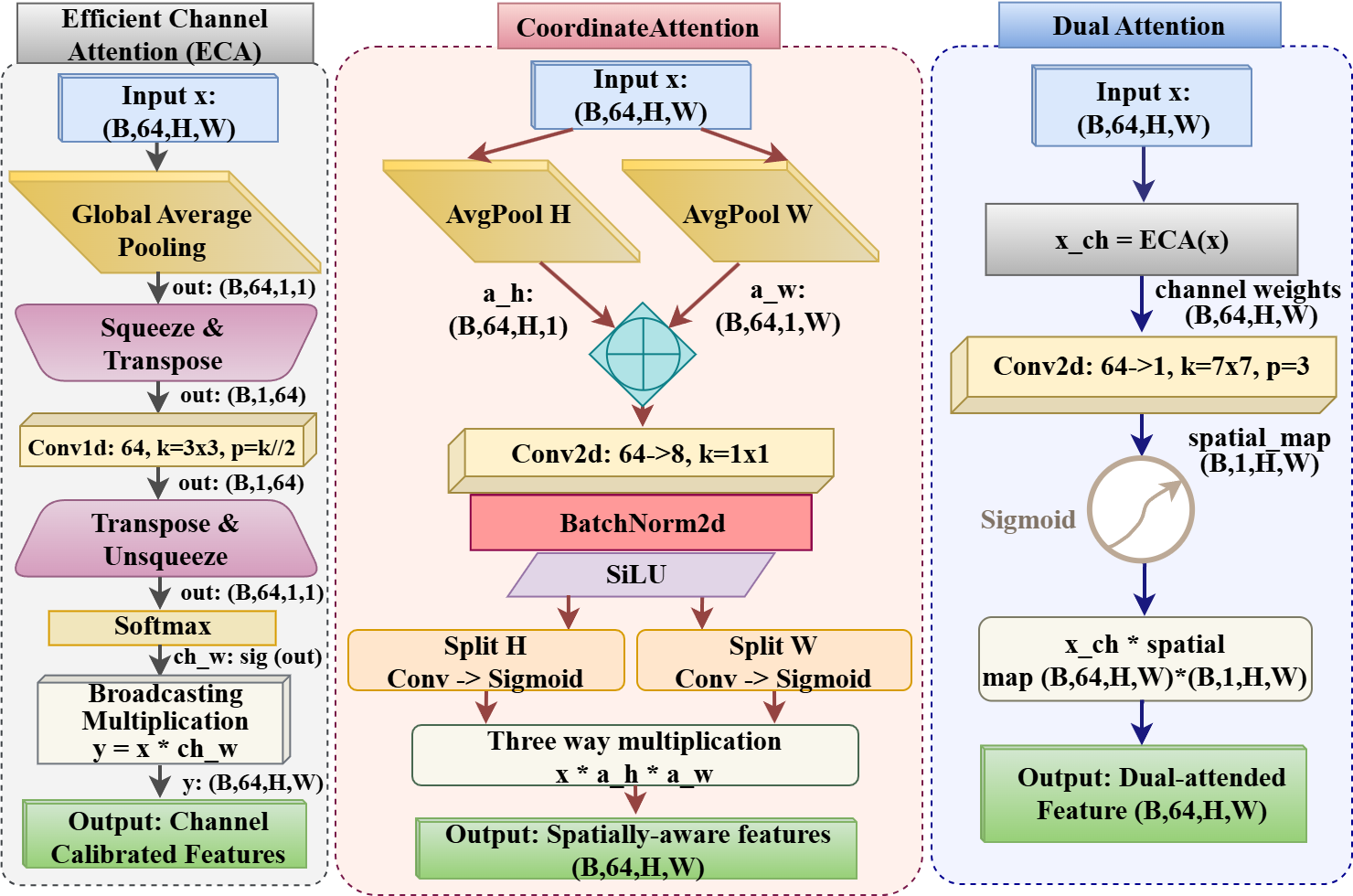}
\caption{Overview of the proposed attention mechanisms integrated in HREM-Net, where ECA emphasizes important channels, Coordinate Attention preserves spatial structure, and Dual Attention refines features for improved segmentation and boundary accuracy.}
\label{fig:atn}
\end{figure}

Coordinate Attention performs separate pooling of input features along the height and width axes, concatenates the resulting descriptors, and passes them through a shared lightweight transformation before splitting them back into height-wise and width-wise attention maps, each generated via a dedicated projection and sigmoid activation. These two maps are then applied multiplicatively to the input, allowing the module to jointly reweight features along both spatial dimensions. Equation \ref{eq25} summarizes this process compactly:

\begin{equation}
    \mathbf{F}_{\text{CA}} 
    = \mathbf{F} 
    \odot \sigma_{\text{sig}}\!\left(\mathrm{Conv}_{h}\!\left(\mathbf{z}\right)\right) 
    \odot \sigma_{\text{sig}}\!\left(\mathrm{Conv}_{w}\!\left(\mathbf{z}\right)\right)
    \label{eq25}
\end{equation}

where
$\mathbf{z} = \phi\!\left(
    \mathrm{Conv}_{1\times1}\!\left(
        \left[\mathrm{Pool}_{h}(\mathbf{F})\,\|\,\mathrm{Pool}_{w}(\mathbf{F})\right]
    \right)
\right)$
is a shared intermediate representation obtained by concatenating the height-wise and width-wise average-pooled descriptors and passing them through a $1{\times}1$ convolution followed by Batch Normalisation and SiLU activation (denoted jointly as $\phi$); $\mathrm{Conv}{h}$ and $\mathrm{Conv}{w}$ are axis-specific $1{\times}1$ projections that decode $\mathbf{z}$ into height and width attention maps, respectively; $\sigma{\text{sig}}(\cdot)$ is the sigmoid activation; and $\odot$ denotes element-wise multiplication. Figure \ref{fig:atn} represents the complete scenario of the attention mechanisms.

\subsection{Cross-Modal Fusion Module}
After extracting features from each modality, an attention-guided dynamic gated fusion strategy is applied to fuse the features, addressing the drawbacks of the noisy feature concatenation method. The encoded hyperspectral and RGB feature maps are, $\mathbf{F}^{h}$ and $\mathbf{F}^{r}$. Since their channel distributions may differ, both are projected into a shared embedding space using $1\times1$ convolutions, defined as $\tilde{\mathbf{F}}^{h} = \phi_{h}(\mathbf{F}^{h})$ and $\tilde{\mathbf{F}}^{r} = \phi_{r}(\mathbf{F}^{r})$, where $\phi_h(\cdot)$ and $\phi_r(\cdot)$ map both features to a common dimension. The projected features are then concatenated and passed through Coordinate Attention to enable cross-modal interaction. This step allows the fused tensor to encode both spectral and spatial cues while maintaining coordinate awareness. 

Further, adaptive gating is employed to dynamically balance the contributions of hyperspectral and RGB feature streams based on global contextual information. A compact descriptor is first obtained through global pooling, which captures the overall scene characteristics. This representation is then passed through a lightweight transformation to produce modality-specific responses, followed by normalization to ensure a relative weighting between the two streams. In this way, the network adaptively emphasizes the more informative modality for a given input, enabling robust feature fusion under varying material and appearance conditions. The final fused representation is obtained by adaptively weighting the projected hyperspectral and RGB features using these gating coefficients. Specifically, the fused feature tensor is computed as $\mathbf{F}_{f} = \alpha_{h}\tilde{\mathbf{F}}^{h} + \alpha_{r}\tilde{\mathbf{F}}^{r}$, where $\alpha_h$ and $\alpha_r$ dynamically control the contribution of each modality for the current input sample.

\begin{algorithm}[ht!]
\caption{Cross-Modal Fusion}
\begin{algorithmic}[1]
\Require HSI features $\mathbf{F}_{HSI} \in \mathbb{R}^{C_h \times H \times W}$, RGB features $\mathbf{F}_{RGB} \in \mathbb{R}^{C_r \times H \times W}$, output dimension $D$
\Ensure Fused features $\mathbf{F}_{out} \in \mathbb{R}^{D \times H \times W}$, gating weights $(\alpha, \beta)$
\If{$\text{shape}(\mathbf{F}_{HSI}) \neq \text{shape}(\mathbf{F}_{RGB})$}
    \State $\mathbf{F}_{RGB} \gets \text{BilinearInterpolate}(\mathbf{F}_{RGB}, (H, W))$ 
\EndIf
\State $\mathbf{F}_h \gets \text{SiLU}(\text{BN}(\text{Conv}_{1 \times 1}(\mathbf{F}_{HSI}, D)))$ 
\State $\mathbf{F}_r \gets \text{SiLU}(\text{BN}(\text{Conv}_{1 \times 1}(\mathbf{F}_{RGB}, D)))$ 
\State $\mathbf{F}_{concat} \gets \text{Concat}([\mathbf{F}_h, \mathbf{F}_r], \text{dim}=1)$ 
\State $\mathbf{F}_{attended} \gets \text{CoordinateAttention}(\mathbf{F}_{concat})$ 
\State $\mathbf{G} \gets \text{GlobalAvgPool}(\mathbf{F}_{attended})$ 
\State $\mathbf{G} \gets \text{SiLU}(\text{Conv}_{1 \times 1}(\mathbf{G}, D))$ 
\State $[\alpha, \beta] \gets \text{Softmax}(\text{Conv}_{1 \times 1}(\mathbf{G}, 2))$ 
\State $\mathbf{F}_{weighted} \gets \alpha \odot \mathbf{F}_h + \beta \odot \mathbf{F}_r$ 
\State $\mathbf{F}_{final} \gets \text{Concat}([\mathbf{F}_{weighted}, \mathbf{F}_{attended}], \text{dim}=1)$ 
\State $\mathbf{F}_{out} \gets \text{DualAttention}(\text{SiLU}(\text{BN}(\text{Conv}_{3 \times 3}(\mathbf{F}_{final}, D))))$ 
\State \Return $\mathbf{F}_{out}, (\alpha, \beta)$
\end{algorithmic}
\label{alg:cross_modal_fusion}
\end{algorithm}

This dynamic weighting mechanism is an important part of the proposed framework, as it enables scene-dependent selection of modality information. This is necessary when lighting conditions or spectral contrast change across different acquisition setups. To further reduce cross-modal noise and strengthen the final fused representation, the fused feature map is concatenated with the attended tensor and passed through a refinement block based on convolution and DA, as defined in equation \ref{eq29}:
\begin{equation}
\mathbf{F}_{\text{final}} = \Psi\left([\mathbf{F}_{f}; \mathbf{F}_{c}]\right)
\label{eq29}
\end{equation}
where $\Psi(\cdot)$ denotes the refinement module. This final step ensures that the fused features are both spatially consistent and material-discriminative before decoding. Algorithm \ref{alg:cross_modal_fusion} describes the proposed Cross-Modal Fusion module, detailing feature alignment, attention-guided feature integration, and adaptive modality weighting through dynamic gating to produce fused representations.

\subsection{Decoder and Segmentation Head}

After feature fusion, the decoder reconstructs full-resolution segmentation logits using a gradual upsampling process. Since accurate boundary reconstruction is important for electrolyzer components, overly aggressive upsampling is avoided. Instead, a multi-stage decoder is employed, where attention-based refinement is applied at each stage. Let the fused feature representation be denoted as $\mathbf{F}_{\text{final}} \in \mathbb{R}^{D_f \times h \times w}$, where $D_f$ represents the number of channels and $(h, w)$ denotes the spatial resolution before decoding. The decoder is initialized as $\mathbf{F}_{u}^{(0)} = \mathbf{F}_{\text{final}}$. At each decoder stage, bilinear upsampling is followed by convolution and normalization, as defined in equation \ref{eq30}:
\begin{equation}
\mathbf{F}_{u}^{(k+1)} = \sigma\left(\mathrm{BN}\left(\mathbf{W}_{u}^{(k)} * \mathrm{Up}(\mathbf{F}_{u}^{(k)})\right)\right)
\label{eq30}
\end{equation}
where $\mathrm{Up}(\cdot)$ indicates bilinear upsampling with a scaling factor of 2, and $\mathbf{W}_{u}^{(k)}$ represents the convolution parameters at the $k$-th decoder level. Dual Attention (DA) is applied at each decoder stage to improve boundary clarity and reduce small incorrect predictions, especially in confusing regions. The final segmentation head converts the refined decoder features into class-wise prediction maps at the original image resolution. These maps show how confident the model is for each class at every pixel. The final segmentation result is obtained by assigning each pixel to the class with the highest score, producing a dense map where each location is labeled as one of the target electrolyzer component classes. To support stable training and improve learning for underrepresented classes, the decoder also produces an auxiliary prediction at an intermediate resolution. This auxiliary output is upsampled to the original image size and used as an additional supervision signal. This design helps improve gradient flow and reduces overfitting to dominant classes.

\subsection{Multi-Component Loss Function}
The dataset shows a strongly imbalanced class distribution, where several important material classes cover far fewer pixels than the main structural regions. In such cases, training with standard cross-entropy loss alone often favors majority classes and causes consistent under-segmentation of rare materials. To overcome this issue, a composite loss function is adopted, integrating PolyLoss \citep{leng2022polyloss}, Tversky loss \citep{salehi2017tversky}, and auxiliary deep supervision \citep{lee2015deeply}, which combines confidence-aware classification and overlap-based optimization for class imbalance. The overall loss function is defined in equation \ref{eq33}:
\begin{equation}
\mathcal{L} = w_{\text{poly}}\mathcal{L}_{\text{poly}} + w_{\text{tv}}\mathcal{L}_{\text{tv}} + w_{\text{aux}}\mathcal{L}_{\text{aux}}
\label{eq33}
\end{equation}
where the loss weights are fixed as PolyLoss $w_{\text{poly}}=1.0$, Tversky Loss $w_{\text{tv}}=0.5$, and Auxiliary Loss $w_{\text{aux}}=0.4$.

\textbf{PolyLoss:} $\mathcal{L}_{\text{poly}}$ augments cross-entropy by adding a polynomial penalty term that emphasizes hard examples through increased contribution from low-confidence predictions as defined in equation \ref{eq34}:
\begin{equation}
\mathcal{L}_{\text{poly}} = \frac{1}{N}\sum_{i=1}^{N}\left[\mathcal{L}_{\text{CE}}^{(i)} + \epsilon(1 - p_t^{(i)})\right]
\label{eq34}
\end{equation}
where $N$ denotes the total number of pixels used to compute the loss. The term $\mathcal{L}_{\text{CE}}^{(i)}$ represents the class-weighted cross-entropy loss for pixel $i$, and $p_t^{(i)}$ denotes the softmax probability assigned to the correct class for pixel $i$. The parameter $\epsilon$ controls the strength of the polynomial term and is set to $\epsilon = 1.0$. The factor $(1 - p_t^{(i)})$ assigns a higher penalty to low-confidence predictions, which improves training stability when the model encounters unclear material boundaries.

\textbf{Tversky Loss:} To explicitly address class imbalance, we include Tversky loss \citep{salehi2017tversky}, which generalizes the Dice coefficient by allowing asymmetric penalties on false positives and false negatives:
\begin{equation}
\mathcal{L}_{\text{tv}} = 1 - \frac{1}{N_b C}\sum_{b=1}^{N_b}\sum_{c=1}^{C}\frac{TP_{bc} + \delta}{TP_{bc} + \alpha FP_{bc} + \beta FN_{bc} + \delta}
\label{eq35}
\end{equation}
Here in equation \ref{eq35}, $N_b$ is the batch size and $C$ is the number of classes. 
For each sample $b$ and class $c$, the overlap statistics are computed across all spatial locations $(i,j)$. Specifically, the true positives are obtained as $TP_{bc} = \sum_{i,j} \hat{y}_{bc}^{(i,j)} \cdot y_{bc}^{(i,j)}$, the false positives are computed as $FP_{bc} = \sum_{i,j} \hat{y}_{bc}^{(i,j)} \cdot (1 - y_{bc}^{(i,j)})$, and the false negatives are calculated as $FN_{bc} = \sum_{i,j} (1 - \hat{y}_{bc}^{(i,j)}) \cdot y_{bc}^{(i,j)}$. Here $\hat{y}_{bc}^{(i,j)}$ is the softmax probability for class $c$ at spatial location $(i,j)$ in sample $b$, and $y_{bc}^{(i,j)}$ is the corresponding one-hot encoded ground truth. The sums run over all spatial locations $(i,j)$. The parameters $\alpha=0.7$ and $\beta=0.3$ control the relative penalty on false positives and false negatives, respectively, highlighting recall over precision to reduce missed detections of minority classes. The smoothing constant $\delta=1.0$ ensures numerical stability.

\textbf{Auxiliary Loss:} The auxiliary loss term $\mathcal{L}_{\text{aux}}$ is computed using standard (unweighted) cross-entropy on intermediate decoder predictions:
\begin{equation}
\mathcal{L}_{\text{aux}} = -\frac{1}{N}\sum_{i,j}\sum_{c=1}^{C} y_{c}^{(i,j)} \log \hat{y}_{\text{aux},c}^{(i,j)},
\label{eq39}
\end{equation}
In equation \ref{eq39} $\hat{y}_{\text{aux},c}^{(i,j)}$ denotes the softmax probability from the auxiliary segmentation head for class $c$ at spatial location $(i,j)$. This deep supervision encourages early decoder layers to learn discriminative representations and improves convergence stability across diverse scenes. Together, these loss components provide a robust optimization objective that maintains boundary precision while significantly reducing bias toward dominant classes under severe material imbalance.

\section{Experimental Details}\label{sec4}
This section describes the dataset preparation process, preprocessing pipeline, and experimental setup for training and evaluating our HREM-Net model.

\subsection{Dataset}
In our study, we used two public datasets, Electrolyzers-HSI \citep{arbash2025electrolyzers} and PCB-Vision \citep{arbash2024pcb}.

\textbf{Electrolyzers-HSI:} The Electrolyzers-HSI dataset \citep{arbash2025electrolyzers} comprises 55 co-registered high-resolution RGB images and HSI data cubes spanning the 400–2500 nm spectral range. The dataset includes shredded electrolyzer fragments from three main material types: High-Temperature Electrolyzer (HTEL) ceramics, Ni-mesh, and interconnector steel plates, collected from both new and end-of-life samples. These materials form five classes: Mesh, Steel-Black, Steel-Grey, HTEL Anode, and HTEL Cathode. Each scene is scanned on both sides to ensure full spectral coverage and is organized into multi-class settings. 

\textbf{PCB-Vision:} The PCB-Vision dataset \citep{arbash2024pcb} is a multiscene RGB-hyperspectral benchmark comprising 53 high-resolution RGB images paired with hyperspectral data cubes, collected from 53 distinct PCBs in an industrial conveyor belt setup. Each image provides pixel-level annotations for three PCB component classes: integrated circuits (ICs), capacitors, and connectors. The dataset captures diverse board layouts and component compositions across multiple scenes, making it a challenging benchmark for evaluating segmentation models. Table \ref{tab:dataset_summary} summarizes the key characteristics of the Electrolyzers-HSI and PCB-Vision dataset.

\begin{table}[ht!]
\scriptsize
\centering
\caption{Dataset Summary: Electrolyzers-HSI and PCB-Vision.}
\label{tab:dataset_summary}
\begin{tabular}{p{1.8cm} p{3.6cm} p{2.4cm}}
\hline
\textbf{Attribute} & \textbf{Electrolyzers-HSI} & \textbf{PCB-Vision} \\
\hline
Dataset Size        & 55 co-registered scenes & 53 co-registered scenes \\
Modalities          & RGB + HSI & RGB + HSI \\
Spatial Resolution  & RGB: 4096 px/line; HSI: 384 px/line & RGB: 4000$\times$3000 px; HSI: 1024 px/line \\
Spectral Range      & 400--2500 nm (VNIR-SWIR) & 400--1000 nm (VNIR) \\
Material Classes    & 5 electrolyzer material types & 3 PCB component types \\
Class Labels        & Mesh, Steel-Black, Steel-Grey, HTEL-Anode, HTEL-Cathode & IC, Capacitor, Connectors \\
Annotation Type     & Pixel-wise segmentation masks & Pixel-wise segmentation masks \\
Application         & E-waste recycling (electrolyzers) & E-waste recycling (PCBs) \\
\hline
\end{tabular}
\end{table}

\subsection{Data Preprocessing} 
\textbf{HSI Preprocessing.} The hyperspectral data is processed through sequential steps to ensure reliable quality and efficient computation. First, radiometric calibration converts raw digital numbers into reflectance values. Then, each spectral band is normalized using percentile-based scaling with the 1st and 99th percentiles as bounds, where values outside this range are clipped, and the rest are scaled to [0, 1], reducing outliers while preserving material-specific information. To reduce dimensionality and computation, spectral binning is applied by removing the first 50 and last 40 noisy bands, reducing 450 bands to 360, and then averaging neighboring bands in groups of 10 to obtain 36 bands. This approach balances preserving important spectral details with improved training stability and faster convergence.

\textbf{RGB Preprocessing.} RGB images are loaded in standard 8-bit format and converted from the BGR color space to RGB. Pixel values are then normalized to the range [0, 1] by dividing by 255. To maintain spatial consistency with the HSI data, RGB images are resized to match the spatial resolution of the HSI using bilinear interpolation when required. All RGB images are finally converted to 32-bit floating-point format to ensure compatibility with the subsequent neural network processing steps. 

\textbf{Data Augmentation.} To improve model robustness and generalization, data augmentation is applied during training as a part of the training process. The augmentation includes horizontal and vertical flipping, each applied with a probability of 0.5, and is performed consistently on the HSI data, RGB images, and ground truth masks to preserve spatial alignment. No rotation or color-based augmentation is used for HSI data in order to protect the original spectral information. All augmented samples are generated dynamically during training, which increases the effective dataset size without requiring additional storage.

\subsection{Implementation Setup}
This study is conducted using an AMD Ryzen 7700 8-core Central Processing Unit (CPU) with 32 GB of system memory. For graphical processing, an NVIDIA GPU with CUDA support and at least 8 GB of video RAM (VRAM). The experiments are performed on an Ubuntu 24 operating system, and Jupyter Notebook is used as the development environment. In this study, five-fold cross-validation is adopted for both datasets due to their limited size and to enable a more reliable and less biased performance estimation.

\section{Result Analysis}\label{sec5}
This section presents the experimental outcomes of the proposed HREM-Net. The performance is evaluated through pixel-wise and object-wise classification metrics, including per-class IoU, accuracy, confusion matrices, and overall segmentation quality. Additionally, ablation studies demonstrate the contribution of individual architectural components, and comparisons with SOTA and current methods highlight the effectiveness of the proposed multimodal fusion approach.

\subsection{Proposed Model's Performance}
\begin{table*}[ht!]
\scriptsize
\centering
\caption{Five-fold validation performance of HREM-Net on the Electrolyzers-HSI dataset.}
\label{tab:overall_metrics}
\begin{tabular}{lccccc}
\toprule
\textbf{Fold} & \textbf{mIoU} & \textbf{Pixel Accuracy (\%)} & \textbf{Mean Class Accuracy (\%)} & \textbf{FWIoU} & \textbf{mAP@0.5} \\
\midrule
1 & 0.8424 & 98.79 & 93.50 & 0.9778 & 0.7823 \\
2 & 0.9004 & 99.15 & 96.52 & 0.9837 & 0.7848 \\
3 & 0.8177 & 98.49 & 93.65 & 0.9729 & 0.7749 \\
4 & 0.6676 & 97.69 & 79.10 & 0.9627 & 0.6007 \\
5 & 0.8772 & 98.96 & 95.46 & 0.9803 & 0.7524 \\
\midrule
\textbf{Overall} & \textbf{0.8211} & \textbf{98.62} & \textbf{91.66} & \textbf{0.9755} & \textbf{0.7391} \\
\bottomrule
\end{tabular}
\end{table*}

\begin{figure}[ht!]
\centering
\includegraphics[scale=0.53]{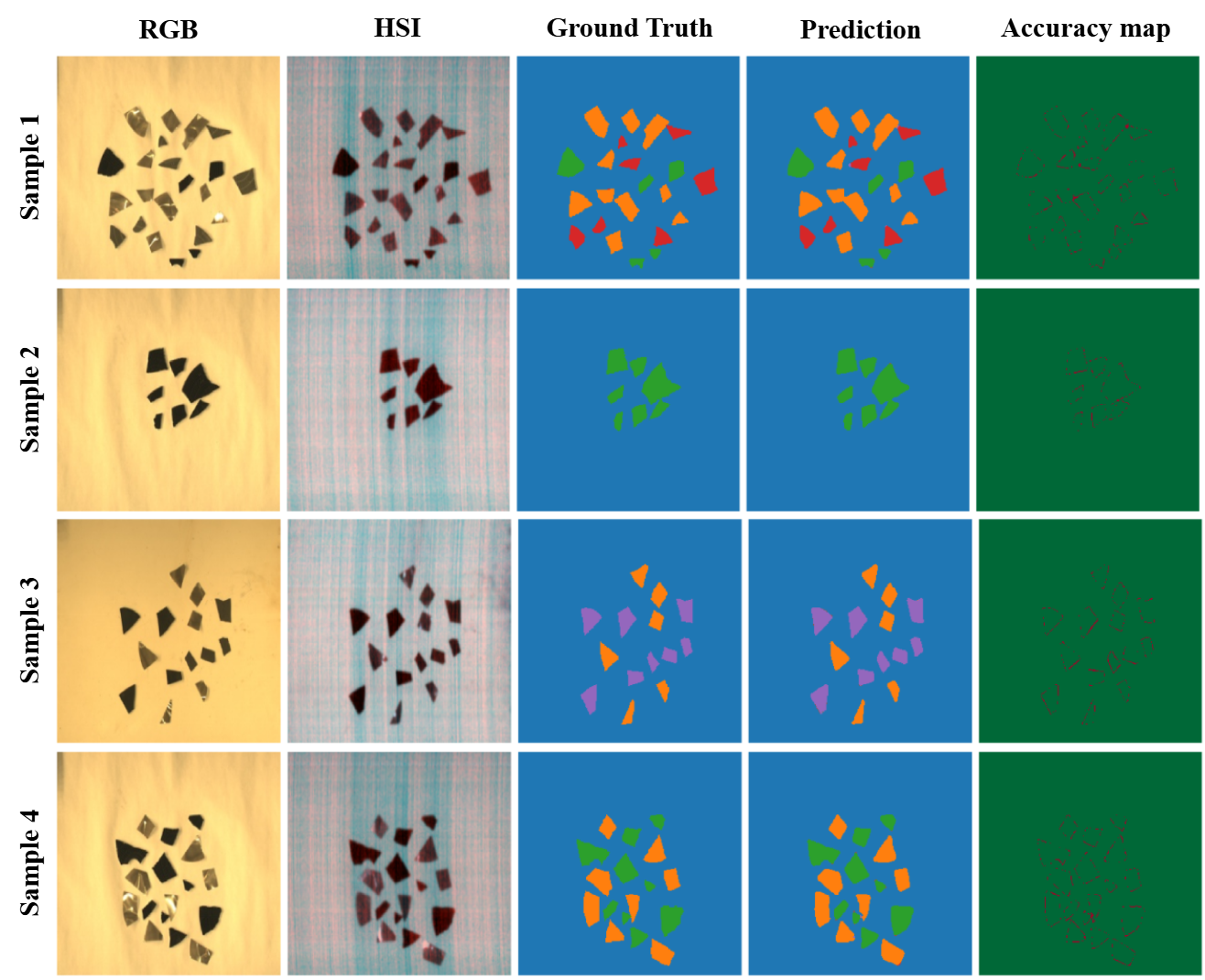}
\caption{Qualitative segmentation results of HREM-Net on representative samples from the Electrolyzers-HSI.}
\label{fig:sample}
\end{figure}

The overall performance across the five-fold validation is summarized in Table \ref{tab:overall_metrics}. HREM-Net achieves an average pixel accuracy of 98.62\%, showing that most pixels are correctly classified across different data splits. The mean class accuracy reaches 91.66\%, which indicates that the model performs reasonably well across the six classes, including the background class, rather than depending only on dominant categories. The average mIoU of 0.8211 shows strong overlap between the predicted segmentation maps and the ground truth masks, while the mAP@0.5 value of 73.91\% further supports reliable region-level prediction. In addition, the frequency weighted IoU FWIoU score of 0.9755 shows that the model remains stable even under severe class imbalance. A noticeable drop is observed in Fold 4, where the mIoU decreases to 0.6676, and the mean class accuracy drops to 79.10\%. The small size and imbalanced nature of the Electrolyzers-HSI dataset mainly cause this reduction. Since each validation fold contains only a limited number of samples, a few difficult or unevenly distributed images with uneven classes can strongly affect the fold-level result. Figure \ref{fig:sample} shows representative segmentation examples, demonstrating that HREM-Net can accurately identify different material classes while preserving object boundaries and fine structural details.

\begin{figure*}[ht!]
\centering
\includegraphics[scale=0.245]{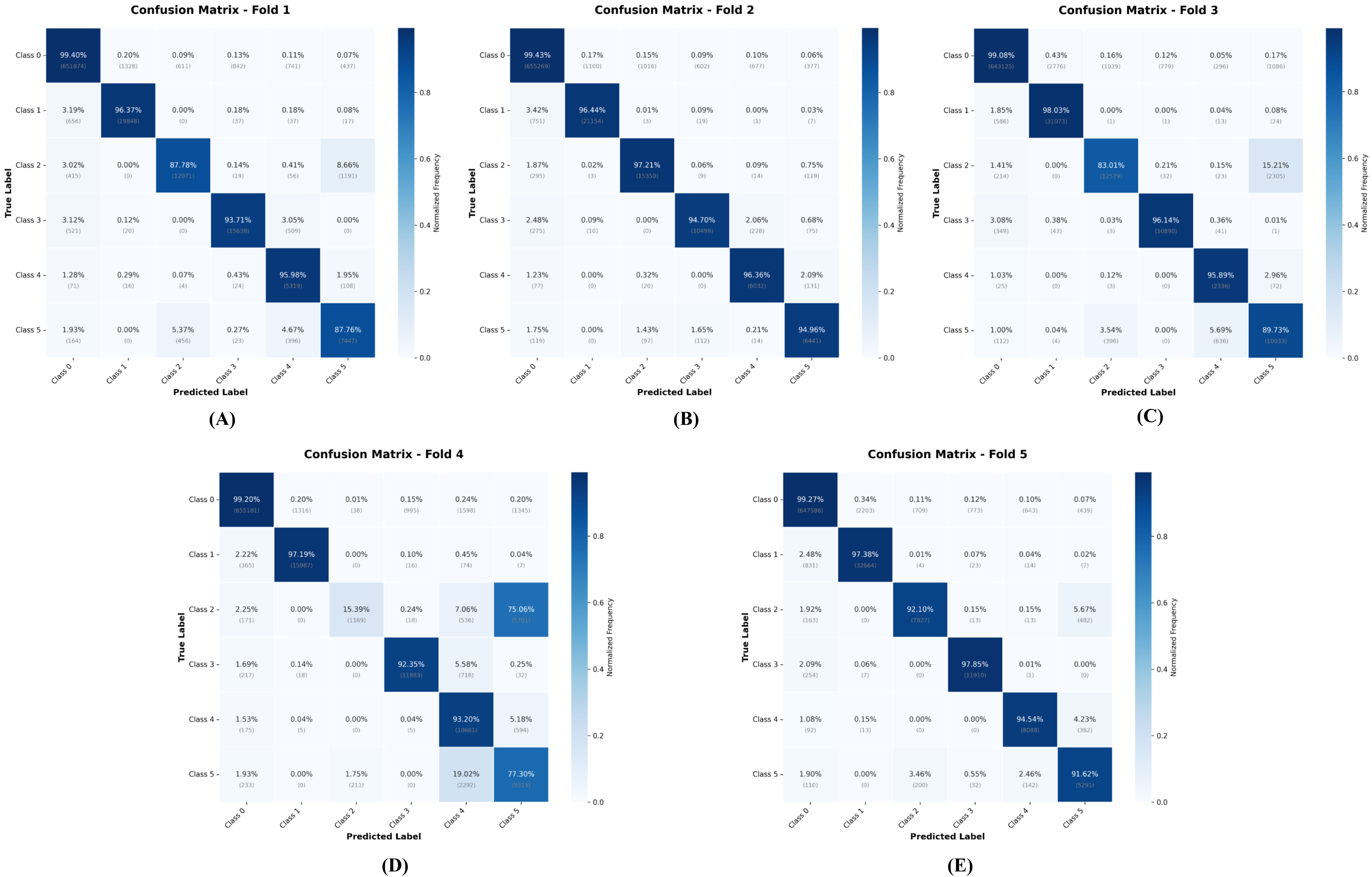}
\caption{Confusion matrices of HREM-Net across five-fold validation on the Electrolyzers-HSI dataset.}
\label{fig:confusion_matrix}
\end{figure*}

\begin{figure*}[ht!]
\centering
\includegraphics[scale=0.35]{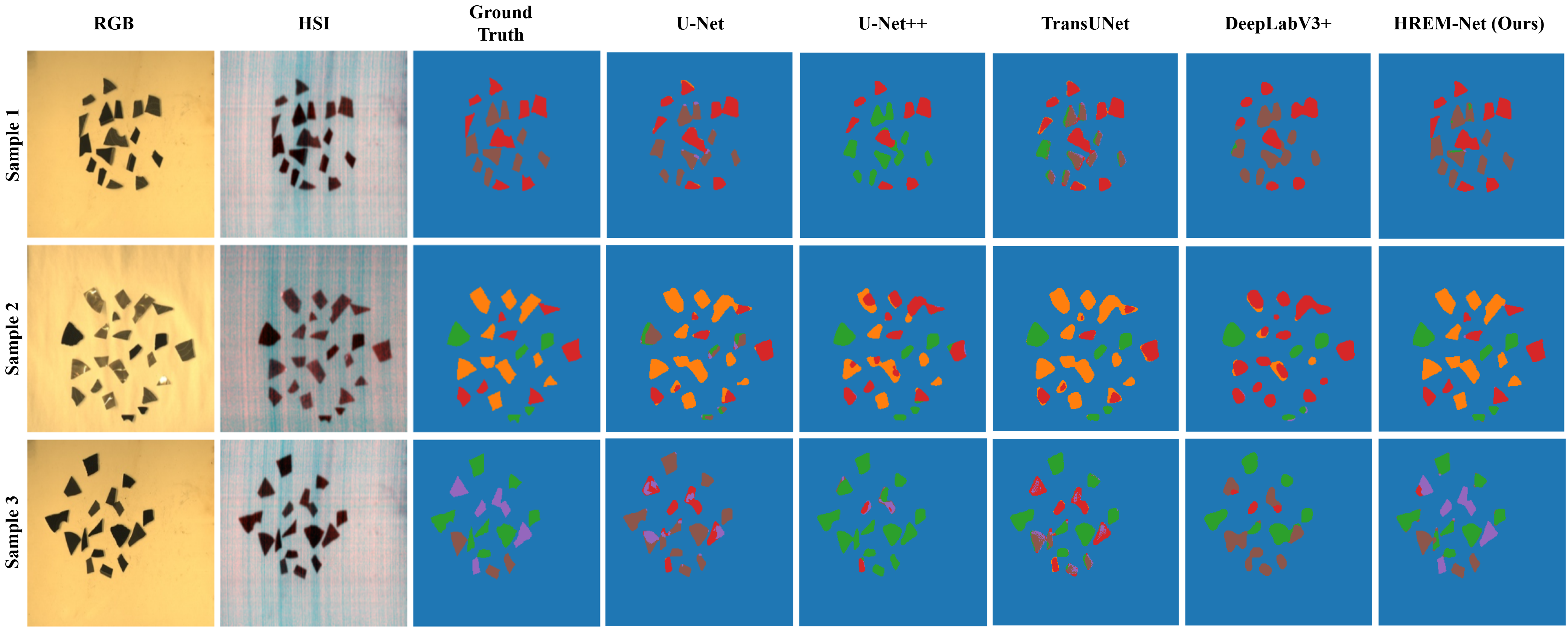}
\caption{Qualitative comparison between HREM-Net and SOTA segmentation models on Electrolyzers-HSI. Each row shows the input images (RGB \& HSI), ground truth, and predictions from different models. Proposed HREM-Net produces more complete regions and cleaner boundaries.}
\label{fig:sota_visual_comparison}
\end{figure*}

To further examine class-level performance, the confusion matrices for all five folds are presented in Figure \ref{fig:confusion_matrix}. Overall, a strong concentration along the diagonal can be observed in most folds, indicating that the majority of samples are correctly classified. This confirms that HREM-Net maintains reliable class-level prediction across different data splits. However, noticeable variations appear across folds, particularly in Fold 4. Figure \ref{fig:confusion_matrix}(D) shows a noticeable drop in class-level performance, mainly due to strong confusion between Class 2 and Class 5, where many samples from Class 2 are incorrectly predicted. The reduced performance is primarily due to the small validation set, uneven sample distribution, and strong class imbalance, which make the problem more severe in this fold, as even a few difficult or less representative samples can significantly affect the results. Despite this fold-level variation, the remaining folds show consistently strong results, confirming the overall robustness of HREM-Net.

\subsection{Comparison with State of the Art}

We compare our proposed HREM-Net with several well-known SOTA segmentation models, including U-Net, U-Net++, DeepLabV3+, and TransUNet on Electrolyzers-HSI. These models represent strong baselines from both convolutional and hybrid encoder families and are commonly used in material and object segmentation tasks.

\begin{table}[ht!]
\scriptsize
\centering
\caption{Comparison of HREM-Net with SOTA segmentation models on the Electrolyzers-HSI dataset.}
\label{tab:sota_comparison}
\begin{tabular}{
p{2.5cm}
>{\centering\arraybackslash}p{1.5cm}
>{\centering\arraybackslash}p{2cm}
>{\centering\arraybackslash}p{1cm}
}
\hline
\textbf{Model} & \textbf{Pixel Accuracy (\%)} & \textbf{Mean Class Accuracy (\%)} & \textbf{mIoU} \\
\hline
U-Net \citep{ronneberger2015u} & 95.71 & 56.11 & 0.4260 \\
U-Net++ \citep{zhou2018unet++} & 94.61 & 54.30 & 0.3685 \\
DeepLabV3+ \citep{chen2018encoder} & 93.94 & 48.07 & 0.3409 \\
TransUNet \citep{chen2021transunet} & 95.68 & 55.46 & 0.4284 \\
\midrule
\textbf{HREM-Net (Ours)} & \textbf{98.62} & \textbf{91.66} & \textbf{0.8211} \\
\hline
\end{tabular}
\end{table}

To ensure a fair comparison, all models are evaluated on the Electrolyzers-HSI dataset using the same evaluation metrics. Table \ref{tab:sota_comparison} presents the quantitative results. The proposed HREM-Net consistently outperforms all competing SOTA methods across every metric. It achieves the highest pixel accuracy of 98.62\%, a mean class accuracy of 91.66\%, and a mIoU of 0.8211. In comparison, the existing models report noticeably lower mIoU values, which suggests weaker segmentation performance at the region level.

Although some SOTA models achieve acceptable pixel accuracy, their mean class accuracy and mIoU remain relatively low, showing difficulty in segmenting smaller classes and visually similar materials. In contrast, HREM-Net provides more balanced performance across all classes by learning both local details and global context, which improves class separation and reduces confusion. Figure \ref{fig:sota_visual_comparison} shows that HREM-Net produces more complete regions, sharper boundaries, and fewer misclassified fragments, while other methods often miss small components or generate fragmented results in complex scenes.

\subsection{Comparison with Current Studies}

\begin{table*}[ht!]
\centering
\small
\caption{Comparison of HREM-Net with existing segmentation studies. Reported results are taken directly from the papers.}
\scriptsize
\begin{tabular}{
>{\raggedright\arraybackslash}p{3.5cm}
>{\centering\arraybackslash}p{4.5cm}
>{\centering\arraybackslash}p{5cm}
>{\centering\arraybackslash}p{4cm}
}
\midrule
\textbf{Study} & \textbf{Model} & \textbf{Dataset} & \textbf{Performance} \\
\midrule
\citep{wang2025cross} & Two-Stage Transformer Ensemble & Custom Construction Recycling & Accuracy: 90.30\%, mIoU: 0.8236 \\
\citep{rahmatulloh2025wasteinnet} & WasteInNet & WasteIn & mAP@0.5: 0.868 \\
\citep{lu2022using} & Updated DeepLabv3+ & Custom Construction Waste & mIoU: 0.562 \\
\citep{yudin2024hierarchical} & H-YC & WaRP & mIoU: 0.6588 \\
\citep{iliushina2025data} & YOLOv8 with data-centric pipeline & Custom MSW & mAP@0.5: 0.7 \\
\citep{tang2022deep} & U-ResNet & PEMFC membrane electrode assembly & Accuracy: 92\%, mIoU: 0.64 \\
\citep{tan2026image} & U-Net Inspired 3D CNN & Solid Oxide Fuel Cell & Overall Accuracy $>$ 90\% \\
\citep{picon2024hyperspectral} & VGG-Style U-Net & Tecnalia WEEE Hyperspectral Dataset & Accuracy: 95\%, mIoU: 0.74 \\
\citep{jafar2026ensemble} & EL-4 & ZeroWaste-f & IoU: 0.8306 \\
\citep{senanayake2025automated} & VLPart & Custom Electro-Construction Waste & mAP@50: 0.396 \\
\midrule
\textbf{Ours} & \textbf{HREM-Net} & \textbf{Electrolyzers-HSI Dataset} & \textbf{Accuracy: 98.62}\%, \textbf{mIoU: 0.8211}, \textbf{mAP@0.5: 0.7391} \\
\midrule
\end{tabular}
\label{tab:comparison_current_studies}
\end{table*}

Earlier works mainly focused on single-modality segmentation or standard encoder–decoder structures. For example, U-ResNet \citep{tang2022deep} achieved 92\% accuracy and an mIoU of 0.64 on PEMFC membrane electrode assembly data. Similarly, the U-Net inspired 3D CNN in \citep{tan2026image} reported overall accuracy above 90\% for anode segmentation tasks. While these results show good classification ability, the reported mIoU values indicate that boundary precision and class-level consistency still have room for improvement. In hyperspectral waste analysis, the VGG-style U-Net \citep{picon2024hyperspectral} achieved 95\% accuracy and 0.74 mIoU on the Tecnalia WEEE dataset. Although the accuracy is comparable, the mIoU remains lower than our method. Frameworks such as WasteInNet \citep{rahmatulloh2025wasteinnet} and YOLOv8 with data-centric design \citep{iliushina2025data} reported mAP@0.5 scores of 0.868 and 0.7, respectively. Other waste-related segmentation studies, such as the updated DeepLabv3+ \citep{lu2022using} and the H-YC framework \citep{yudin2024hierarchical}, reported mIoU scores of 0.562 and 0.6588, respectively, showing moderate segmentation performance in construction and recycling datasets. In contrast, our proposed HREM-Net achieves 98.62\% pixel accuracy and an improved mIoU of 0.8211 on the Electrolyzers-HSI dataset. These results indicate that the model separates material classes more clearly and detects object boundaries more precisely during segmentation. 

Table \ref{tab:comparison_current_studies} presents a comparison between our proposed HREM-Net and several representative studies related to industrial material and waste segmentation. Since these works use different datasets and object categories, the comparison should be understood as a general performance overview rather than a direct benchmark.

\subsection{Cross Dataset Validation}
\begin{figure}[ht!]
\centering
\includegraphics[scale=0.25]{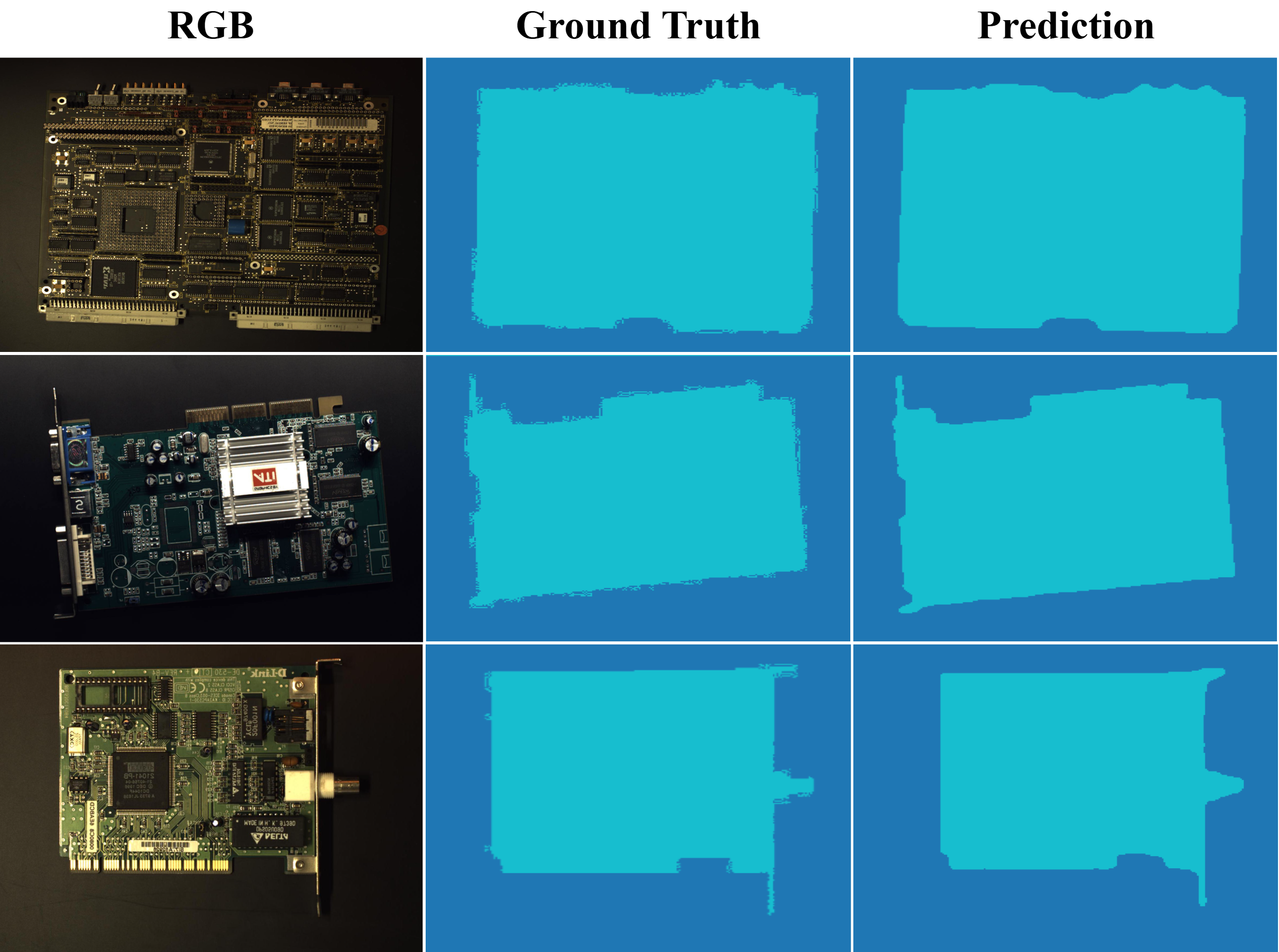}
\caption{Qualitative cross-dataset validation performance of HREM-Net on the PCB-Vision dataset.}
\label{fig:pcbvision}
\end{figure}

To evaluate the robustness of HREM-Net, we evaluated it on the PCB-Vision dataset, which contains HSI images, RGB images, and ground truths of printed circuit boards. Figure \ref{fig:pcbvision} shows qualitative results, where the predicted masks closely match the ground truth across various PCB samples, confirming the model's effectiveness.

Quantitative results in Table \ref{tab:cross_dataset_results} show that HREM-Net maintains strong and consistent performance on the PCB-Vision dataset. The model achieves an overall Pixel Accuracy of 96.88\% and a Mean Class Accuracy of 96.91\%, along with an mIoU of 0.9396 and an FWIoU of 0.9397. Across individual folds, the performance remains stable, with mIoU values ranging from 0.9124 to 0.9546 and Pixel Accuracy varying between 95.42\% and 97.68\%. These results indicate that HREM-Net demonstrates strong robustness and consistent segmentation performance across different datasets.

\begin{table}[ht!]
\scriptsize
\centering
\caption{Performance of HREM-Net on the PCB-Vision dataset.} 
\label{tab:cross_dataset_results}
\begin{tabular}{
>{\centering\arraybackslash}p{1cm}
>{\centering\arraybackslash}p{1cm}
>{\centering\arraybackslash}p{1.8cm}
>{\centering\arraybackslash}p{1.8cm}
>{\centering\arraybackslash}p{1.2cm}
}
\toprule
\textbf{Fold} & \textbf{mIoU} & \textbf{Pixel Accuracy (\%)} & \textbf{Mean Class Accuracy (\%)} & \textbf{FWIoU} \\
\midrule
1 & 0.9407 & 96.95 & 96.93 & 0.9408 \\
2 & 0.9510 & 97.49 & 97.57 & 0.9511 \\
3 & 0.9546 & 97.68 & 97.66 & 0.9547 \\
4 & 0.9124 & 95.42 & 95.53 & 0.9124 \\
5 & 0.9392 & 96.87 & 96.82 & 0.9393 \\
\midrule
\textbf{Overall} & \textbf{0.9396} & \textbf{96.88} & \textbf{96.91} & \textbf{0.9397} \\
\bottomrule
\end{tabular}
\end{table}

\subsection{Ablation Study}
To better understand the contribution of each component in the proposed HREM-Net architecture, we conduct a detailed ablation study. The analysis focuses on four key elements: MBCov, ECA, ASPP, and the Tversky loss function.

\begin{table}[ht!]
\scriptsize
\centering
\caption{Ablation study of HREM-Net showing the impact of different architectural components.}
\label{tab:ablation_study}
\begin{tabular}{
>{\centering\arraybackslash}p{1.2cm}
>{\centering\arraybackslash}p{0.8cm}
>{\centering\arraybackslash}p{0.9cm}
>{\centering\arraybackslash}p{1.2cm}
>{\centering\arraybackslash}p{0.8cm}
>{\centering\arraybackslash}p{1.5cm}
}
\hline
\textbf{MBCov} & \textbf{ECA} & \textbf{ASPP} & \textbf{Tversky Loss} & \textbf{mIoU} & \textbf{Pixel Acc (\%)} \\
\hline
\cmark & \xmark & \xmark & \xmark & 0.24 & 89.89 \\
\xmark & \cmark & \xmark & \xmark & 0.63 & 96.95 \\
\xmark & \xmark & \cmark & \xmark & 0.62 & 96.89 \\
\cmark & \cmark & \xmark & \xmark & 0.36 & 92.44 \\
\cmark & \xmark & \cmark & \xmark & 0.52 & 96.46 \\
\xmark & \cmark & \cmark & \xmark & 0.54 & 96.12 \\
\cmark & \cmark & \cmark & \cmark & \textbf{0.82} & \textbf{98.62} \\
\hline
\end{tabular}
\end{table}

Table \ref{tab:ablation_study} presents the results of different architectural configurations. When only one module is used, the segmentation performance remains limited. For example, the configuration that includes only MBConv achieves a low mIoU of 0.24, showing that relying only on spatial feature diversity is not sufficient for accurate segmentation. Using ECA or ASPP individually leads to noticeable improvements in pixel accuracy and moderate increases in mIoU, highlighting the importance of channel attention and multi-scale contextual information. After that, we evaluate configurations that combine two modules. Although these combinations further increase pixel accuracy compared to single-module settings, the mIoU remains relatively low. This indicates that higher pixel-level accuracy does not necessarily translate into better region-level segmentation. The results suggest that partial combinations are still unable to effectively address both class imbalance and boundary precision at the same time.

The final configuration integrates all components, including MBCov, ECA, ASPP, and the Tversky loss. This complete model achieves the best performance, with a mIoU of 0.82 and a pixel accuracy of 98.62\%. The large improvement compared to all other variants confirms that each module provides a correlative contribution. More specifically, combining multi-scale feature extraction, attention-based feature refinement, and a class-aware loss function leads to more accurate and stable segmentation outcomes. Overall, this ablation study shows that the strong performance of HREM-Net arises from the effective integration of all architectural elements, rather than from any single component.

\section{Discussion}\label{sec6}
Autonomous hydrogen production monitoring is a crucial step in hydrogen production, and a data-driven intelligent electrolyzer systems can facilitate this production. We aim to contribute in these aspects by developing a robust AI framework, HREM-Net, integrating several advanced modules jointly to improve accuracy and robustness for electrolyzer material segmentation. To address the drawbacks of existing studies, including single data modality~\citep{rahmatulloh2025wasteinnet, lu2022using}, poor segmentation performance \citep{yudin2024hierarchical, tang2022deep}, we design the innovative HREM-Net, a dual-branch deep learning architecture, which can process the HSI and RGB images simultaneously. The innovative ECA and coordinate attention mechanism capture wavelength-dependent material properties, while MBConv blocks with SE and ASPP can extract texture and structural features. The novel cross-modal fusion module refines the features, enabling the model to distinguish visually similar materials that differ in their spectral signatures. The combination of PolyLoss \citep{leng2022polyloss}, Tversky loss \citep{salehi2017tversky}, and auxiliary deep supervision \citep{lee2015deeply} further improves the training process by addressing class imbalance and enhancing boundary-level learning. Each of these components contributes to the overall segmentation performance in a clearly measurable manner.

The HREM-Net model demonstrates strong segmentation performance on the Electrolyzers-HSI \citep{arbash2025electrolyzers} dataset across all evaluated metrics. The model achieved a pixel accuracy of 98.62\%, a mean class accuracy of 91.66\%, and an mIoU of 0.8211, confirming its ability to segment both dominant and minority material classes reliably. The high pixel accuracy reflects overall efficiency, while the mIoU score indicates precise region-level segmentation with clear class boundaries. As shown in Table \ref{tab:ablation_study}, the ablation study confirms the progressive contribution of each proposed component. The combined use of MBConv, ECA, ASPP, and Tversky Loss explains the superior performance of HREM-Net, as its strength comes from the integration of all components rather than a single module.
Moreover, the model outperforms the SOTA models \citep{ronneberger2015u, zhou2018unet++, chen2018encoder}, highlighting its segmentation superiority. In addition, the generalization capability of HREM-Net is further confirmed by its strong performance on the PCB-Vision \citep{arbash2024pcb} cross-dataset evaluation, obtaining pixel accuracy of 96.88\%, a mean accuracy of 96.91\%, a mean IoU of 0.9396, and a FWIoU of 0.9397 on the dataset. These overall results confirm the generalizability, consistency, and robustness of the proposed model.

Although HREM-Net shows strong performance, some limitations remain. The model depends on well-aligned hyperspectral and RGB data, so misalignment, noise, or poor calibration can reduce accuracy, especially at boundaries. In real industrial environments, factors such as surface contamination, corrosion, dust, and varying illumination may change both spectral responses and visual appearance, which could lead to performance degradation. Future work will focus on improving robustness to real-world noise and environmental variability, incorporating more diverse data, and extending the framework to support a larger number of material categories.

\section{Conclusion}\label{sec7}
This work presents HREM-Net, a dual-branch deep learning framework that combines hyperspectral and RGB imaging for reliable semantic segmentation of electrolyzer components. The Spectral Encoder uses spectral compression with ECA and coordinate attention to process high-dimensional hyperspectral data efficiently while preserving material-specific features. In parallel, the RGB Encoder applies MBConv blocks with SE and ASPP to capture fine spatial details and multi-scale context from RGB images. By integrating both modalities through a cross-modal fusion module with dynamic gating and coordinate attention, the framework achieves strong segmentation performance on the Electrolyzers-HSI dataset, reaching 98.62\% pixel accuracy, 91.66\% mean class accuracy, and mIoU of 0.8211, clearly outperforming existing and SOTA methods.

Extensive result analysis shows that our AI-based visual analysis can be an effective tool for monitoring and operating assessment of intelligent electrolyzers. The proposed HREM-Net will contribute to improving reliability, reducing maintenance requirements, and improving efficiency of hydrogen production processes by enabling automated interpretation of electrical chemical system behaviour. Future work will focus on real industrial deployment settings, extending material diversity, and evaluating robustness to support sustainable hydrogen recycling and circular economy goals.

\section*{Statements and Declarations}

\subsection*{Funding} This study does not include any external funding.
\subsection*{Ethical Approval and Consent to Participate} Not applicable
\subsection*{Consent for Publication} 
All the authors have reviewed all the versions of the manuscript and approved the final version for submission.
\subsection*{Competing interests:} The authors state that they have no known financial conflicts of interest or personal relationships that could have influenced the work presented in this paper.
\subsection*{Availability of Data and Materials} In this study, we have used two publicly available datasets, \href{https://rodare.hzdr.de/record/3668}{Electrolyzers-HSI} \citep{arbash2025electrolyzers} and \href{https://rodare.hzdr.de/record/2704}{PCB-Vision} \citep{arbash2024pcb}.


\subsection*{CRediT Author Statement}
\textbf{Conceptualization:} Wasimul Karim, Nur Mohammad Fahad, Abdul Hasib Siddique; \textbf{Investigation:} Wasimul Karim, Nur Mohammad Fahad; \textbf{Methodology: }Wasimul Karim, Nur Mohammad Fahad, Sami Azam; \textbf{Validation:} Wasimul Karim, Nur Mohammad Fahad, Abdul Hasib Siddique, Md Rafiqul Islam, Hooman Mehdizadeh-Rad, Asif Karim, Sami Azam; \textbf{Visualization:} Wasimul Karim; \textbf{Writing – original draft:} Wasimul Karim, Nur Mohammad Fahad, Abdul Hasib Siddique; \textbf{Writing – review \& editing:} Wasimul Karim, Nur Mohammad Fahad, Abdul Hasib Siddique, Md Rafiqul Islam, Hooman Mehdizadeh-Rad, Asif Karim, Sami Azam; \textbf{Project administration:} Nur Mohammad Fahad, Abdul Hasib Siddique, Sami Azam; \textbf{Formal Analysis:} Md Rafiqul Islam, Hooman Mehdizadeh-Rad, Asif Karim; \textbf{Resources:} Wasimul Karim; \textbf{Supervision:} Nur Mohammad Fahad, Sami Azam.

\end{document}